\documentclass[a4paper,fleqn]{elsarticle}

\usepackage{amsmath}
\usepackage{amssymb}
\usepackage{newtxtext,newtxmath}
\usepackage{booktabs}
\usepackage{graphicx}
\usepackage{multirow}
\usepackage{array}
\usepackage{longtable}
\usepackage{pdflscape}
\usepackage{siunitx}
\usepackage{url}
\usepackage[hidelinks]{hyperref}
\usepackage{tabularx}

\begin{document}

\begin{frontmatter}



\title{Beyond aggregate scores: Deployment-aware and non-compensatory
benchmarking of vision-based eye-state recognition models for driver monitoring}




\author[aff1]{Ruben Dario Florez-Zela\corref{cor1}}
\cortext[cor1]{Corresponding author}
\ead{rflorezz@unsa.edu.pe}
\affiliation[aff1]{
    organization={Universidad Nacional de San Agustin de Arequipa (UNSA)},
    city={Arequipa},
    postcode={04001},
    country={Peru}
}

\begin{abstract}
Model selection for safety-relevant visual recognition is often based on clean aggregate performance, although robustness, transfer, embedded latency, and explanation faithfulness may produce different preferences. This study presents a Human-Centered Benchmarking Framework (HCBF) that separates multidimensional evidence from non-compensatory operational eligibility. Six compact convolutional and transformer-oriented eye-state recognition models were evaluated using a subject-disjoint MRL Eye protocol, deterministic image corruptions, zero-shot transfer and participant-safe target-domain training with out-of-fold evaluation on RT-BENE, TensorRT FP32 inference on an NVIDIA Jetson Nano, and black-box RISE faithfulness. Clean MRL Macro-F1 ranged from 0.9566 to 0.9794, whereas zero-shot RT-BENE Macro-F1 ranged from 0.2066 to 0.7771. Matched target-domain effects varied from -0.0406 to 0.4807 and redistributed the two directional errors differently across architectures. Only MobileNetV3-Large and ShuffleNetV2 met the 33.333-ms binocular-pair latency deadline, while neither passed the predefined safety-related screen. The remaining four models failed both requirements, yielding an empty eligible set. Normalized deletion AUC ranged from 0.5826 to 0.9113, while normalized insertion coverage varied from 17.6\% to 88.6\%. Model ordering changed across clean prediction, corruption robustness, transfer, deployment, faithfulness, and historical score sensitivity. These findings show that relative ranking, multidimensional preference, and operational eligibility are distinct decisions. Deployment-aware benchmarking should preserve directional failures and uncertainty and should allow no model to be selected when mandatory requirements are unmet.
\end{abstract}


\end{frontmatter}



\section{Introduction}
\label{sec:introduction}

Vision-based driver monitoring provides non-contact evidence about visual behavior and readiness to remain engaged with the driving task \cite{qu2024driver,albadawi2022review}. Within this broader setting, eye-state recognition supports the estimation of blink dynamics, prolonged closure, and ocular indicators such as PERCLOS. These indicators are associated with reduced alertness but do not constitute a complete diagnosis of drowsiness when used alone \cite{kolus2024eye,abe2023perclos}. Eye-state recognition is therefore treated here as an enabling visual function rather than an end-to-end drowsiness detector.

Most classifiers are initially judged under clean, within-dataset conditions. That evidence is necessary but incomplete because in-cabin vision is affected by subject variability, illumination, sensor properties, blur, occlusion, and head movement. Natural datasets such as RT-BENE differ substantially from controlled near-infrared eye-image collections \cite{cortacero2019rtbene}. Models with similar held-out performance may consequently implement different decision functions and fail differently after deployment \cite{damour2022underspecification}. Aggregate metrics can further conceal miscalibration \cite{guo2017calibration} and asymmetric errors whose operational consequences are not equivalent.

Robustness and transfer must also be separated. Standard corruption benchmarks show marked deterioration under noise, blur, and digital distortions \cite{hendrycks2019corruptions}, whereas synthetic robustness transfers only partially to naturally occurring shifts \cite{taori2020naturalshift}. Domain-generalization research likewise shows that conclusions depend on the datasets, architectures, and model-selection protocol \cite{zhou2023domain,wang2023domain}. A credible benchmark should therefore combine controlled perturbations with external-domain evaluation and determine whether limited target-domain training has consistent effects across architectures.

Deployment introduces another independent source of divergence. Parameter count and floating-point operations do not fully determine latency because memory access, operator implementation, compilation, and hardware scheduling shape realized execution. ShuffleNetV2 and hardware-aware mobile design made this distinction explicit \cite{ma2018shufflenetv2,howard2019mobilenetv3}. Accordingly, target-device latency and its upper tail should be measured directly rather than inferred from nominal complexity or desktop timing.

Explanation evidence is similarly conditional. Saliency maps can appear plausible while remaining weakly related to the learned function \cite{adebayo2018sanity,hooker2019benchmark}. RISE provides a common black-box procedure for heterogeneous architectures because it requires only model inputs and outputs \cite{petsiuk2018rise}. Its deletion and insertion metrics offer quantitative faithfulness evidence, but their validity depends on the perturbation baseline, stochastic masks, and sample coverage.

The unresolved problem is thus not the absence of individual metrics, but their integration into model selection. Reporting dimensions independently leaves ranking stability unclear, while a compensatory aggregate score can allow strong performance in one dimension to offset a critical failure in another. A model may rank first among the available alternatives and still exceed a latency deadline or violate a safety-related error requirement. Relative ranking and operational eligibility are different decisions, and the latter must permit an empty eligible set.

To address this gap, a Human-Centered Benchmarking Framework (HCBF) is developed for vision-based eye-state recognition. Here, \emph{human-centered} denotes an evaluation design that preserves asymmetric errors, subject-level independence, acquisition changes, deployment constraints, and transparent evidence about model behavior; it does not denote a human-subject evaluation of explanation interfaces. Six compact architectures are examined through clean discrimination and calibration, deterministic corruptions, zero-shot transfer, participant-safe target-domain training, compiled embedded inference, black-box faithfulness, and sensitivity of a historical aggregate score.

The contributions are fourfold:
\begin{itemize}
    \item A unified, subject-disjoint framework integrates predictive,     robustness, transfer, deployment, and faithfulness evidence without     forcing a current universal score.
    \item Directional failures and participant-level uncertainty are retained across controlled corruptions and external-domain evaluation.
    \item Compiled target-hardware measurements are linked to predefined non-compensatory gates, explicitly allowing the eligible set to remain empty.
    \item A black-box RISE audit and a global sensitivity analysis of the historical HCS show that favorable clean or aggregate ranking does not guarantee operational admissibility.
\end{itemize}

Section~\ref{sec:related_work} reviews the relevant literature. Sections~\ref{sec:framework} and \ref{sec:experimental_setup} present HCBF and the experimental protocol. Section~\ref{sec:results} reports the results, which are interpreted in Section~\ref{sec:discussion}. Limitations and conclusions follow in Sections~\ref{sec:limitations} and \ref{sec:conclusions}.
\section{Related work}
\label{sec:related_work}

\subsection{Eye-state recognition and driver-monitoring datasets}

Eye-state recognition has progressed from geometric and hand-crafted appearance descriptors to compact deep models. Song et al. \cite{song2014eyes} combined multi-scale oriented-gradient descriptors and introduced Closed Eyes in the Wild, highlighting pose, illumination, and occlusion variability. MRL Eye provides approximately 85,000 annotated near-infrared eye images from 37 participants and supports controlled subject-disjoint evaluation \cite{fusek2018pupil}. RT-BENE instead targets blink estimation in natural environments and provides temporally paired eye regions with substantially different acquisition conditions \cite{cortacero2019rtbene}. Broader resources such as DMD include face, body, hand, RGB, infrared, and depth views for system-level driver-monitoring research \cite{ortega2021dmd}. The present study retains the narrower ocular task so that reliability dimensions can be evaluated under a consistent prediction unit; its conclusions do not extend to complete drowsiness diagnosis.

\subsection{Calibration, robustness, and domain shift}

High discrimination does not imply calibrated confidence \cite{guo2017calibration}, and aggregate performance can conceal asymmetric class failures. ImageNet-C established standardized testing under common corruptions \cite{hendrycks2019corruptions}, while natural-shift studies showed that synthetic robustness is not a substitute for external-domain evaluation \cite{taori2020naturalshift}. WILDS and DomainBed further demonstrated the importance of naturally occurring shifts and consistent model-selection protocols \cite{koh2021wilds,gulrajani2021domainbed}. Surveys of domain generalization emphasize the dependence of conclusions on the assumed training and target distributions \cite{zhou2023domain,wang2023domain}; backbone choice can also materially alter out-of-domain behavior \cite{angarano2024backbones}. HCBF does not introduce a new defense or adaptation algorithm. It applies one deterministic protocol to a fixed architecture portfolio and preserves subject identity as the statistical unit.

\subsection{Efficient architectures and target-hardware evidence}

The six-model portfolio represents distinct efficiency principles. ShuffleNetV2 emphasized memory access and operator fragmentation \cite{ma2018shufflenetv2}; MobileNetV3 combined hardware-aware search with mobile design \cite{howard2019mobilenetv3}; and EfficientNet used compound scaling \cite{tan2019efficientnet}. DeiT improved data-efficient transformer training \cite{touvron2021deit}, EfficientFormer pursued transformer structures at mobile latency \cite{li2022efficientformer}, and RepViT revisited mobile convolution from a transformer-design perspective \cite{wang2024repvit}. Their published latency claims are platform specific. Consequently, parameters and theoretical operations remain descriptors rather than substitutes for measurements after compilation on the intended hardware.

\subsection{Faithfulness and multidimensional reporting}

Grad-CAM is efficient but depends on internal activations \cite{selvaraju2017gradcam}. RISE estimates importance through randomized input masks and is therefore applicable to convolutional and transformer-based models under one black-box procedure \cite{petsiuk2018rise}. Qualitative maps alone are insufficient: sanity checks and remove-and-retrain evaluations have shown that visually plausible explanations may not track learned parameters or predictive relevance \cite{adebayo2018sanity,hooker2019benchmark}. Deletion and insertion also depend on feature removal and may generate out-of-distribution inputs \cite{gomez2022metrics}. Model cards and underspecification research similarly argue for reporting intended use, evaluation conditions, and divergent behavior rather than a single headline metric \cite{mitchell2019modelcards,damour2022underspecification}.

Existing studies therefore provide mature tools for each dimension but normally apply them in separate experimental settings. HCBF addresses the resulting model-selection problem. A descriptive historical score is retained under explicit weights, while current latency and safety-related requirements are enforced through non-compensatory gates. Faithfulness remains separate evidence. This structure distinguishes a favorable relative rank from operational eligibility and permits no candidate to be selected.

The methodological contribution is therefore not a new classifier, corruption operator, or explainer, but a prespecified evaluation and decision architecture that preserves heterogeneous evidence, subject-level uncertainty, and metric validity while linking mandatory requirements to non-compensatory operational eligibility.
\section{Human-Centered Benchmarking Framework}
\label{sec:framework}

\subsection{Evidence architecture}

Let $\mathcal{M}=\{m_1,\ldots,m_K\}$ be the model portfolio. For image $x_i$, model $m$ returns the open-eye probability $p_{m,i}$. The frozen decision is defined in Eq.~\eqref{eq:frozen_decision}.
\begin{equation}
\widehat{y}_{m,i}=\mathbf{1}[p_{m,i}\geq0.5].
\label{eq:frozen_decision}
\end{equation}
with $y_i=1$ for open and $y_i=0$ for the non-open state. The decision threshold and binary open/non-open coding remain unchanged across models and evidence blocks.

\begin{figure}[!h]
    \centering
    \includegraphics[width=\textwidth]{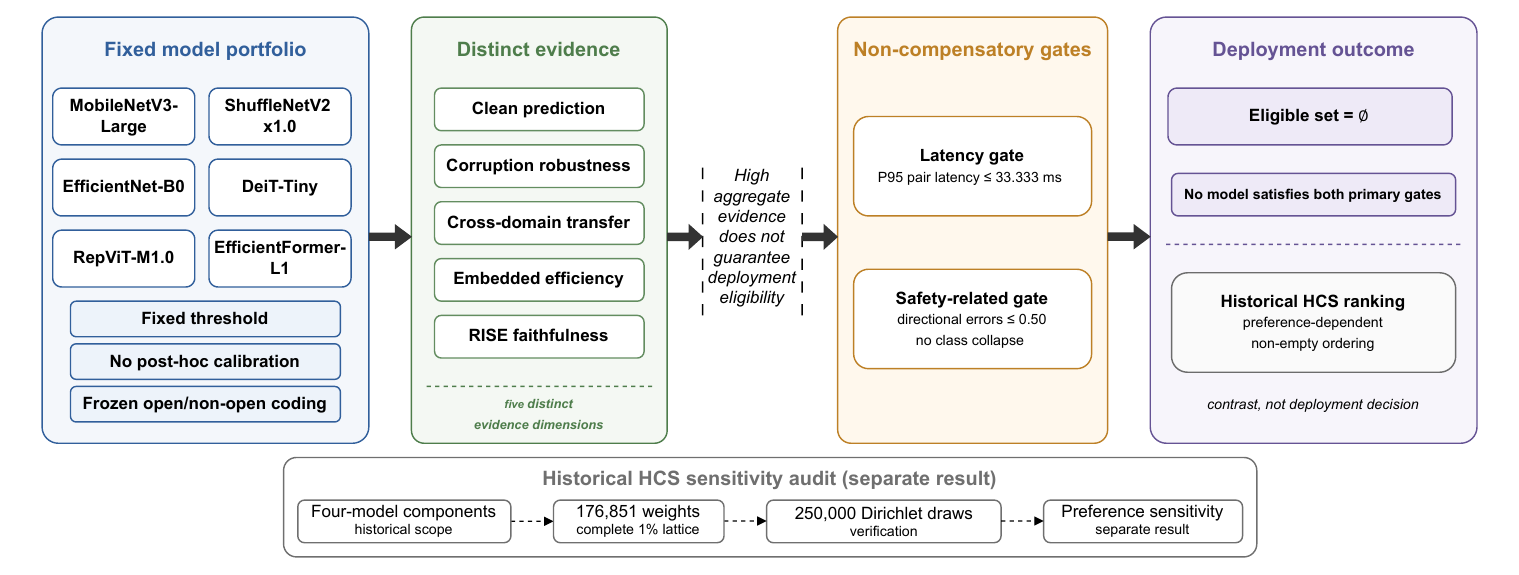}
    \caption{Overview of the HCBF evaluation framework. Six fixed vision architectures are assessed across five distinct evidence dimensions. Operational eligibility is determined separately through non-compensatory latency and safety-related gates, so strong aggregate evidence does not guarantee deployment eligibility. The historical HCS ranking is retained only as a retrospective preference-sensitivity analysis.}
    \label{fig:hcbf_overview}
\end{figure}

As summarized in Fig.~\ref{fig:hcbf_overview}, HCBF retains five current evidence blocks: clean predictive behavior, controlled corruption robustness, cross-domain transfer and target-domain training, embedded inference, and explanation faithfulness. A retrospective lane audits the historical four-model HCS without extending it to the new architectures or inserting RISE evidence.

\subsection{Predictive, robustness, and transfer evidence}

Clean evidence includes accuracy, Macro-F1, balanced accuracy, AUROC, Brier score, ECE, and class-specific metrics. Because the two error directions may have different consequences, HCBF retains the two directional error rates defined in Eq.~\eqref{eq:directional_errors}.
\begin{align}
e_{0\rightarrow1}(m)&=
\frac{\sum_i\mathbf{1}[y_i=0,\widehat y_{m,i}=1]}
{\sum_i\mathbf{1}[y_i=0]}, &
e_{1\rightarrow0}(m)&=
\frac{\sum_i\mathbf{1}[y_i=1,\widehat y_{m,i}=0]}
{\sum_i\mathbf{1}[y_i=1]} .
\label{eq:directional_errors}
\end{align}
They are named closed-to-open and open-to-closed on MRL, and blink-to-open and open-to-blink on RT-BENE.

For metric $q$, the matched corruption and target-domain effects are defined in Eq.~\eqref{eq:matched_effects}.
\begin{align}
\Delta^{(q)}_{m,c}&=q_{m,c}-q_{m,\mathrm{clean}}, &
A^{(q)}_{m,c}&=q^{(\mathrm{OOF})}_{m,c}-q^{(0)}_{m,c}.
\label{eq:matched_effects}
\end{align}
Condition-level results are complemented by family summaries, worst directional errors, and class-collapse indicators. Synthetic corruptions and natural domain shift remain separate because they represent different forms of evidence. Full metric definitions and family aggregation are provided in Supplementary Section S.1.

\subsection{Deployment and non-compensatory eligibility}

Compiled artifacts enter screening only after numerical parity with the frozen reference outputs. The binary parity indicator $g_m^{\mathrm{par}}$ equals one only when the compiled implementation satisfies all prespecified numerical-parity criteria described in Section~\ref{sec4.4}; otherwise, $g_m^{\mathrm{par}}=0$. Let $\ell_m^{95}$ denote the P95 latency for one binocular pair and $\tau_\ell$ the deadline. The latency gate is defined in Eq.~\eqref{eq:latency_gate}.
\begin{equation}
g_m^{\mathrm{lat}}=\mathbf{1}[\ell^{95}_m\leq\tau_\ell].
\label{eq:latency_gate}
\end{equation}
The primary safety-related gate uses a prespecified minimum screening profile. Across every non-clean RT-BENE out-of-fold condition, neither class may collapse and each class must retain a recall of at least 0.50; equivalently, both directional error rates must remain at or below 0.50. This criterion was frozen before Pareto analysis and is used as a comparative benchmarking screen rather than an automotive safety-certification threshold. Numerical parity, latency, and safety-related screening are combined conjunctively as defined in Eq.~\eqref{eq:eligibility}.
\begin{align}
z_m&=g_m^{\mathrm{par}}g_m^{\mathrm{lat}}g_m^{\mathrm{safe}}, &
\mathcal{M}_{\mathrm{elig}}&=\{m\in\mathcal{M}:z_m=1\}.
\label{eq:eligibility}
\end{align}
Strong performance elsewhere cannot compensate for a failed gate, and $\mathcal{M}_{\mathrm{elig}}=\varnothing$ is admissible. Pareto analysis is restricted to eligible models.

\subsection{Architecture-agnostic faithfulness}

For target class $t$, the RISE saliency estimate from randomized masks $M_k$ is defined in Eq.~\eqref{eq:rise_saliency} \cite{petsiuk2018rise}:
\begin{equation}
S_{m,t}(x)=
\frac{1}{K\,\mathbb{E}[M]}
\sum_{k=1}^{K} f_{m,t}(x\odot M_k)M_k .
\label{eq:rise_saliency}
\end{equation}
With target scores $f_0=f_{m,t}(x)$ and $f_b=f_{m,t}(x_b)$, the normalized deletion and insertion areas are defined in Eq.~\eqref{eq:normalized_faithfulness}.
\begin{align}
D_m^{\mathrm{norm}}&=\int_0^1\frac{d_m(u)}{f_0}\,du, &
I_m^{\mathrm{norm}}&=\int_0^1
\frac{i_m(u)-f_b}{f_0-f_b}\,du ,
\quad |f_0-f_b|\geq0.05 .
\label{eq:normalized_faithfulness}
\end{align}
Lower deletion and higher insertion are preferred. Invalid insertion values are not imputed; model-specific and common-valid coverage are reported. Faithfulness is diagnostic and does not enter the operational gates.

\subsection{Historical HCS and uncertainty}

The historical HCS retains accuracy $\alpha_m$, legacy explainability $\epsilon_m$, efficiency $\eta_m$, and robustness $\rho_m$, combined according to Eq.~\eqref{eq:historical_hcs}.
\begin{equation}
\operatorname{HCS}_m(\mathbf w)=
w_\alpha\alpha_m+w_\epsilon\epsilon_m+w_\eta\eta_m+w_\rho\rho_m,
\qquad \sum_j w_j=1.
\label{eq:historical_hcs}
\end{equation}
Its original Safety, Deployment, and Balanced profiles and the full weight simplex are audited only over the historical four-model set. First-place share measures coverage of the preference space, not probability of objective superiority. The legacy normalization and sensitivity formulation are detailed in Supplementary Section S.2.

Images and frames from one participant are correlated. Confidence intervals and paired contrasts therefore resample participant identifiers rather than individual images. Shared resampling weights are used for matched conditions and regimes. This hierarchy preserves the separation among evidence dimensions, mandatory eligibility, restricted Pareto comparison, and historical preference sensitivity.
\section{Experimental setup}
\label{sec:experimental_setup}

\subsection{Datasets and subject partitions}

MRL Eye provides controlled subject-disjoint training and in-domain testing, whereas RT-BENE provides a natural external domain with paired left and right eye regions \cite{fusek2018pupil,cortacero2019rtbene}. Their frozen roles and counts are summarized in Table~\ref{tab:dataset_partitions}.

\begin{table}[!h]
\centering
\caption{Frozen dataset roles and evaluation cohorts. RT-BENE counts use the
binocular frame as the prediction unit; paired eye crops are shown in
parentheses.}
\label{tab:dataset_partitions}
\small
\setlength{\tabcolsep}{4pt}
\begin{tabularx}{\textwidth}{@{}
>{\raggedright\arraybackslash}p{1.55cm}
>{\raggedright\arraybackslash}p{2.15cm}
r
>{\raggedright\arraybackslash}p{2.8cm}
>{\raggedright\arraybackslash}X@{}}
\toprule
Dataset & Cohort & Subjects & Primary units & Class composition \\
\midrule
MRL Eye & Training & 26 & 70,551 images & 31,893 closed; 38,658 open \\
MRL Eye & Validation & 6 & 4,970 images & 2,025 closed; 2,945 open \\
MRL Eye & Test & 5 & 9,377 images & 8,028 closed; 1,349 open \\
RT-BENE & Zero-shot & 14 & 107,350 frames (214,700 eyes) &
4,557 blink; 102,793 open \\
RT-BENE & Matched OOF & 11 & 77,119 frames (154,238 eyes) &
2,892 blink; 74,227 open \\
\bottomrule
\end{tabularx}
\end{table}

MRL contains 84,898 valid images from 37 participants. A fixed seed-42 subject split assigned 26 participants to training, six to validation, and five to testing, with no participant overlap. Labels were closed (0) and open (1). RT-BENE open frames were mapped to 1, blink frames to 0, and annotation disagreements or incomplete eye pairs were excluded. Because the non-open label denotes closed eyes in MRL Eye but blink frames in RT-BENE, the external evaluation represents transfer between related non-open ocular states rather than identical label semantics. The zero-shot cohort contained 107,350 frames from 14 participants. The primary RT-BENE prediction averaged the left- and right-eye probabilities before thresholding.

\subsection{Models and training}

The portfolio comprised MobileNetV3-Large, ShuffleNetV2 x1.0, EfficientNet-B0, DeiT-Tiny, RepViT-M1.0, and EfficientFormer-L1. Exact parameter counts and checkpoint metadata are reported in Supplementary Table S.1. All MRL backbones started from ImageNet-1K weights, received two-class heads, and used $224\times224$ three-channel inputs. Training augmentation included horizontal flips, $\pm10^\circ$ rotation, brightness and contrast jitter, and occasional grayscale conversion.

Cross-entropy and AdamW were used with learning rate $10^{-4}$, weight decay $10^{-2}$, five head-only epochs, and at most 25 joint epochs under cosine annealing. The best validation Macro-F1 checkpoint was retained with patience five, batch size 32, and seed 42. Test data were not used for training, threshold selection, calibration, or checkpoint selection.

The same optimization recipe was used for the participant-safe target-domain runs; each outer-fold model was initialized independently from ImageNet-1K rather than from an MRL checkpoint.

\subsection{Evaluation pathways}

\begin{figure}[!h]
    \centering
    \includegraphics[width=\textwidth]{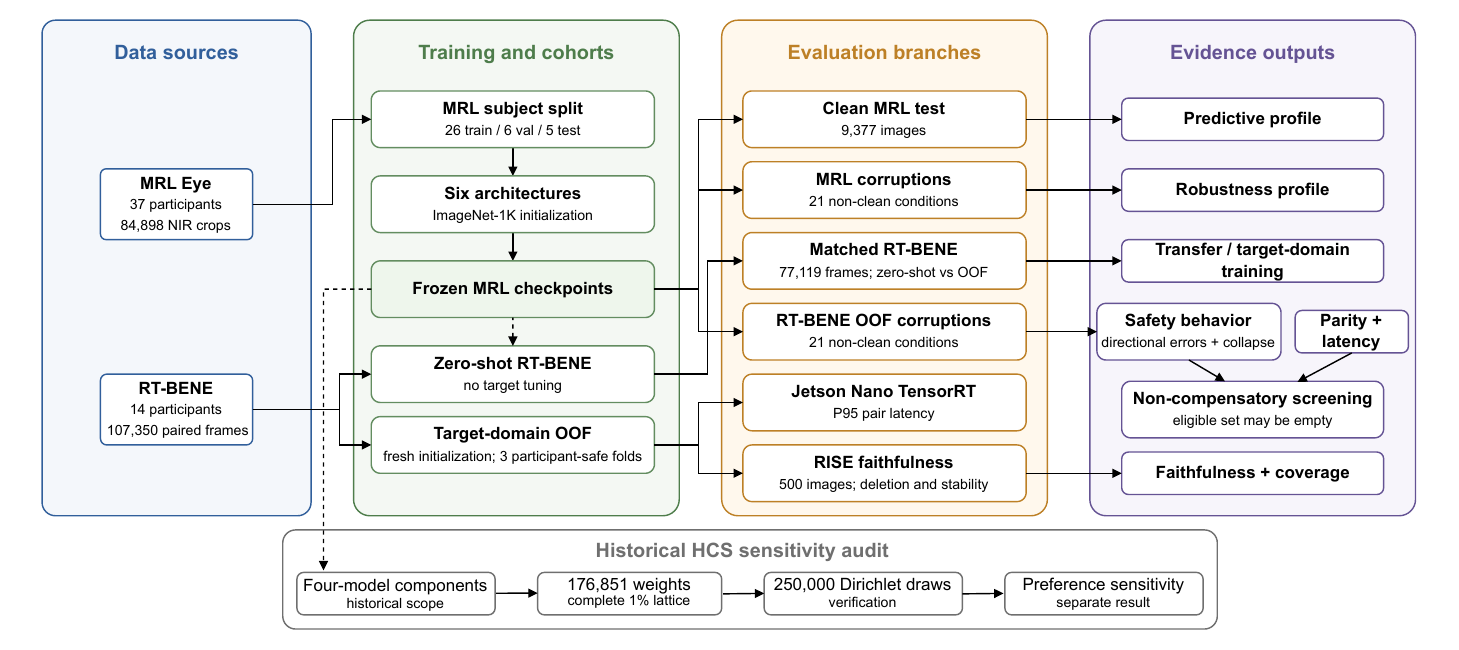}
    \caption{Experimental protocol. MRL Eye provides subject-disjoint training, clean testing, controlled corruption analysis, embedded inference, and RISE. RT-BENE provides zero-shot transfer and participant-safe target-domain out-of-fold evaluation on a matched 11-participant cohort. The same non-clean corruption catalogue is additionally applied to the RT-BENE OOF cohort to obtain the directional-error and class-collapse evidence used by the primary safety-related gate.}
    \label{fig:experimental_protocol}
\end{figure}

Figure~\ref{fig:experimental_protocol} summarizes the complete protocol. MRL provides clean testing, deterministic corruption analysis, embedded benchmarking, and RISE. RT-BENE provides a frozen zero-shot regime and participant-safe target-domain training, followed by matched comparisons on the same frames and participants. The full zero-shot evaluation used all 14 RT-BENE participants (107,350 frames), whereas matched zero-shot versus target-domain comparisons were restricted to the 11-participant cohort (77,119 frames) shared by both regimes. The target-domain out-of-fold cohort is additionally evaluated under the same non-clean corruption catalogue to obtain the directional-error and class-collapse evidence used by the primary safety-related gate.

Clean MRL evaluation used the frozen 9,377-image test set and no post-training calibration. Zero-shot RT-BENE used the MRL checkpoints without target labels, threshold tuning, or model selection. The target-domain regime used fresh ImageNet-1K initialization and three participant-safe outer folds. Three participants were held fixed for validation across folds, whereas the remaining 11 participants were assigned to mutually exclusive outer test folds of four, four, and three participants. All sessions from a given participant remained in the same role within each fold, preventing participant overlap between training, validation, and testing. The resulting 18 model-fold runs produced out-of-fold predictions for 77,119 frames from the 11 outer-fold participants. Zero-shot and target-domain predictions were then aligned by frame for matched comparison.

\subsection{Corruptions, deployment, and RISE}
\label{sec4.4}
The catalogue contained clean plus 21 non-clean conditions: Gaussian noise with standard deviations 10, 25, and 40 on the 8-bit intensity scale, with five deterministic realizations per level; brightness factors 0.5, 0.7, and 1.5; and horizontal motion-blur kernels 5, 11, and 17. The same sample-specific realization was supplied to all models. A three-channel grayscale control was retained as a chroma ablation, not as a simulation of near-infrared sensing.

The corruption catalogue was evaluated on both the clean MRL test partition and the matched RT-BENE out-of-fold cohort. The MRL conditions supported the controlled robustness analysis, whereas the non-clean RT-BENE OOF conditions provided the directional-error and class-collapse evidence used by the primary safety-related gate in Eq.~\eqref{eq:eligibility}. Gray3 was analyzed separately and was not included among the 21 non-clean gate conditions.

Reference timing used an RTX 3060 Laptop GPU. Operational timing used Jetson Nano MAXN, fixed clocks, active cooling, TensorRT FP32, and separate batch-1 and batch-2 engines. Before timing, all 12 TensorRT FP32 engines, comprising six architectures at batch sizes 1 and 2, were required to pass a prespecified numerical-parity gate on 16 frozen real-image MRL Eye validation tensors. TensorRT outputs were compared with the frozen reference outputs, with acceptance requiring a 99th-percentile absolute probability difference no greater than $10^{-4}$, a maximum absolute probability difference no greater than $5\times10^{-4}$, zero class disagreements, and a maximum repeated-run absolute logit difference no greater than $10^{-7}$. All 12 engines satisfied these criteria. After 50 warm-up iterations, five repeats of 200 synchronized invocations were recorded. The primary criterion was batch-2 P95 latency no greater than 33.333 ms, representing 30 binocular pairs/s. Preprocessing and the rest of the driver-monitoring pipeline were excluded.

RISE was evaluated on 500 clean MRL images correctly classified by all six models, balanced by class. Main maps used 4,096 masks on an $8\times8$ grid with activation probability 0.5. A balanced 64-image subset used three independent 2,048-mask banks for stability. Deletion and insertion curves had 101 fractions and a Gaussian-blurred baseline. Normalized deletion was available for all 3,000 model-image pairs; insertion required $|f_0-f_b|\geq0.05$.

\subsection{Statistical analysis}

Participants were the independent resampling units. Clean MRL intervals used 10,000 participant-cluster bootstrap replicates; RT-BENE transfer, target-domain training, and corruption contrasts used 10,000 paired participant-cluster replicates with shared participant weights across compared regimes and conditions. RISE inference used exact two-sided sign-flip tests over the five MRL test participants. Fifteen model pairs were evaluated for each of five prespecified faithfulness metrics, yielding 75 pairwise tests. All $2^5=32$ sign patterns were enumerated, making 0.0625 the minimum attainable two-sided exact $p$-value. Holm adjustment was applied separately within the 15 model-pair comparisons of each metric, and the resulting inference was treated as exploratory. The historical HCS simplex was enumerated at increments of 0.01 (176,851 vectors) and independently checked with 250,000 deterministic Dirichlet draws.

\section{Results}
\label{sec:results}

\subsection{Predictive behavior and transfer}

Table~\ref{tab:predictive_summary} shows that the clean evidence did not identify a model leading every metric. MobileNetV3-Large achieved the highest clean Macro-F1 (0.9794), whereas DeiT-Tiny had the highest balanced accuracy (0.9844) and lowest open-to-closed error (0.0200). ShuffleNetV2 and MobileNetV3-Large had nearly identical closed-to-open errors, but ShuffleNetV2's open-to-closed error was almost three times larger. The zero-shot results in Table~\ref{tab:predictive_summary} use the complete 14-participant RT-BENE cohort.

\begin{table}[!h]
\centering
\caption{Clean MRL and zero-shot RT-BENE predictive profiles. F1 denotes
Macro-F1 and BAcc balanced accuracy. Directional errors are
closed$\rightarrow$open (C$\rightarrow$O), open$\rightarrow$closed
(O$\rightarrow$C), blink$\rightarrow$open (B$\rightarrow$O), and
open$\rightarrow$blink (O$\rightarrow$B). Full calibration results are given
in Supplementary Table S.2.}
\label{tab:predictive_summary}
\scriptsize
\setlength{\tabcolsep}{2.5pt}
\resizebox{\textwidth}{!}{%
\begin{tabular}{lrrrrrrrr}
\toprule
& \multicolumn{4}{c}{Clean MRL} & \multicolumn{4}{c}{Zero-shot RT-BENE} \\
\cmidrule(lr){2-5}\cmidrule(lr){6-9}
Model & F1 & BAcc & C$\rightarrow$O & O$\rightarrow$C &
F1 & BAcc & B$\rightarrow$O & O$\rightarrow$B \\
\midrule
MobileNetV3-Large & 0.9794 & 0.9777 & 0.0052 & 0.0393 &
0.6034 & 0.8774 & 0.0619 & 0.1832 \\
ShuffleNetV2 x1.0 & 0.9566 & 0.9417 & 0.0055 & 0.1112 &
0.2066 & 0.5885 & 0.0105 & 0.8124 \\
EfficientNet-B0 & 0.9625 & 0.9533 & 0.0066 & 0.0867 &
0.5551 & 0.8205 & 0.1255 & 0.2334 \\
DeiT-Tiny & 0.9752 & 0.9844 & 0.0112 & 0.0200 &
0.7771 & 0.9228 & 0.1020 & 0.0523 \\
RepViT-M1.0 & 0.9631 & 0.9538 & 0.0064 & 0.0860 &
0.6387 & 0.8480 & 0.1760 & 0.1281 \\
EfficientFormer-L1 & 0.9780 & 0.9834 & 0.0087 & 0.0245 &
0.6314 & 0.7949 & 0.2965 & 0.1138 \\
\bottomrule
\end{tabular}%
}
\end{table}

The external-domain ordering changed markedly. DeiT-Tiny led zero-shot RT-BENE Macro-F1 (0.7771), while MobileNetV3-Large reached 0.6034. ShuffleNetV2 showed the largest transfer failure, with Macro-F1 0.2066 and an open-to-blink error of 0.8124. Clean calibration also varied despite uniformly high AUROC; full accuracy, AUROC, Brier, and ECE results are reported in Supplementary Table S.2.

\subsection{Controlled corruption robustness}

\begin{figure}[!h]
    \centering
    \includegraphics[width=0.92\textwidth]{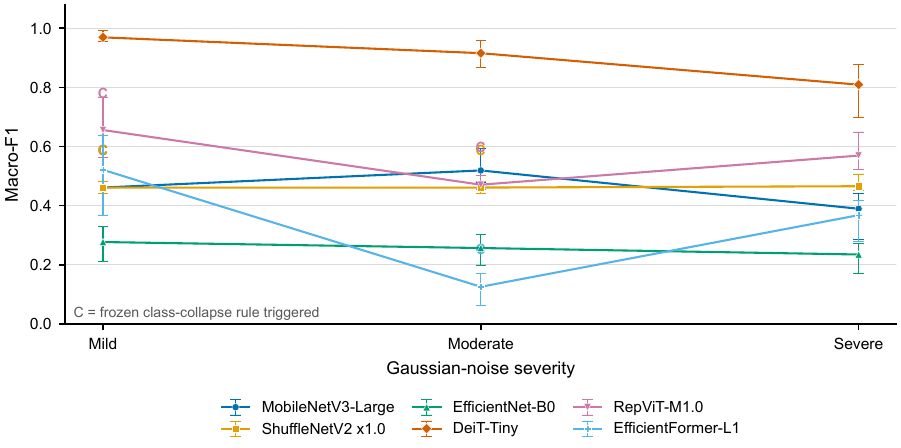}
    \caption{Macro-F1 under Gaussian-noise severity. Points are pooled estimates and bars are 95\% participant-cluster bootstrap intervals.}
    \label{fig:noise_robustness_six_models}
\end{figure}

Gaussian noise produced the strongest and most heterogeneous degradation (Fig.~\ref{fig:noise_robustness_six_models}). DeiT-Tiny declined progressively from Macro-F1 0.9690 under mild noise to 0.8090 under severe noise without collapse. MobileNetV3-Large and ShuffleNetV2 instead collapsed toward closed predictions under mild noise, yielding Macro-F1 0.4612 and open-to-closed error 1. EfficientNet-B0 shifted toward open predictions, while EfficientFormer-L1 approached an all-open decision under moderate noise. RepViT-M1.0 remained strongly biased toward closed decisions.

Blur and brightness were less destructive at the aggregate level but did not preserve the same ordering. Severe-blur Macro-F1 ranged from 0.8683 to 0.9313, and the condition labelled severe brightness ranged from 0.9276 to 0.9825.

The separate Gray3 chroma ablation also changed the matched RT-BENE zero-shot ordering. MobileNetV3-Large increased from 0.5474 to 0.7756 Macro-F1, whereas DeiT-Tiny changed from 0.7443 to 0.7585, shifting the highest Macro-F1 from DeiT-Tiny under RGB to MobileNetV3-Large under Gray3. The effect was strongly architecture dependent and should not be interpreted as a simulation of near-infrared sensing. Complete condition-level results, class-collapse diagnostics, and Gray3 comparisons are reported in Supplementary Tables~S.3--S.5.

\subsection{Matched target-domain effects}

For the matched target-domain analysis, the zero-shot baseline was recomputed on the same 11-participant, 77,119-frame cohort used by the target-domain out-of-fold regime. These matched zero-shot values therefore differ from the 14-participant zero-shot results reported in Table~\ref{tab:predictive_summary}.

\begin{figure}[!h]
    \centering
    \includegraphics[width=0.86\textwidth]{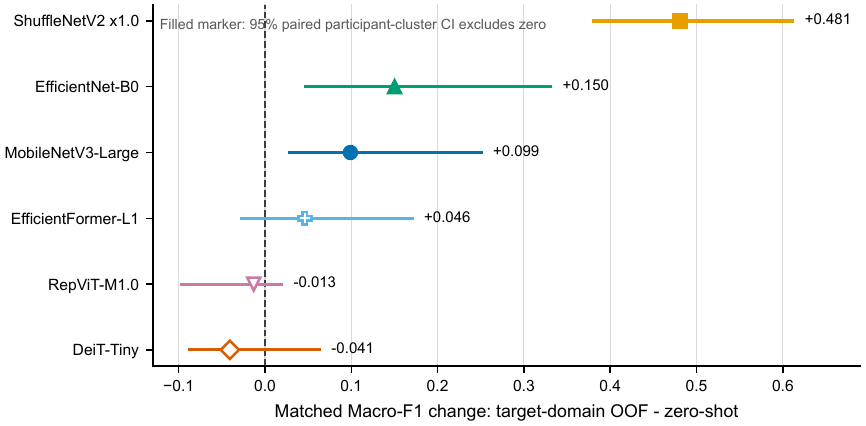}
    \caption{Matched RT-BENE Macro-F1 change after participant-safe target-domain training. Points show out-of-fold minus zero-shot effects; horizontal bars are 95\% paired participant-cluster bootstrap intervals. Filled markers indicate intervals excluding zero.}
    \label{fig:adaptation_macro_f1_effects}
\end{figure}

On the same 77,119 RT-BENE frames, target-domain training improved Macro-F1 most for ShuffleNetV2 (0.4807), EfficientNet-B0 (0.1502), and MobileNetV3-Large (0.0990), with paired intervals excluding zero (Fig.~\ref{fig:adaptation_macro_f1_effects}). EfficientFormer-L1 had a small positive point estimate with an interval spanning zero, while RepViT-M1.0 and DeiT-Tiny had negative point estimates with intervals spanning zero.

The gains were not directionally uniform. ShuffleNetV2 reduced open-to-blink error by 0.6964 while increasing blink-to-open error by 0.1068; MobileNetV3-Large and EfficientNet-B0 showed the same trade-off at smaller magnitudes. Detailed paired effects are given in Supplementary Table S.6.

\subsection{Embedded efficiency and eligibility}

Only ShuffleNetV2 and MobileNetV3-Large met the 33.333-ms P95 deadline (Table~\ref{tab:operational_summary}). The remaining pair latencies ranged from 46.9355 to 64.1434 ms. Desktop and Jetson rankings differed: DeiT-Tiny was fastest on the RTX reference device but ranked fifth for binocular-pair latency on Jetson, with Spearman rank association 0.314. The complete transfer from parameter count to desktop and embedded latency rankings is shown in Supplementary Fig.~S.1.

\begin{table}[!h]
\centering
\caption{Primary Jetson Nano and safety-related screening outcomes. Worst
recall is the minimum class recall across frozen non-clean RT-BENE
out-of-fold conditions.}
\label{tab:operational_summary}
\scriptsize
\setlength{\tabcolsep}{4pt}
\begin{tabularx}{\textwidth}{@{}Xrrrrccc@{}}
\toprule
Model & P95 pair (ms) & Worst recall & Collapses & Latency & Safety & Eligible \\
\midrule
ShuffleNetV2 x1.0 & 12.7055 & 0.0000 & 10 & Yes & No & No \\
MobileNetV3-Large & 23.5928 & 0.1823 & 0 & Yes & No & No \\
EfficientNet-B0 & 46.9355 & 0.1618 & 0 & No & No & No \\
EfficientFormer-L1 & 50.8545 & 0.0060 & 5 & No & No & No \\
DeiT-Tiny & 52.6500 & 0.3053 & 0 & No & No & No \\
RepViT-M1.0 & 64.1434 & 0.0000 & 15 & No & No & No \\
\bottomrule
\end{tabularx}
\end{table}

\begin{figure}[!h]
    \centering
    \includegraphics[width=0.88\textwidth]{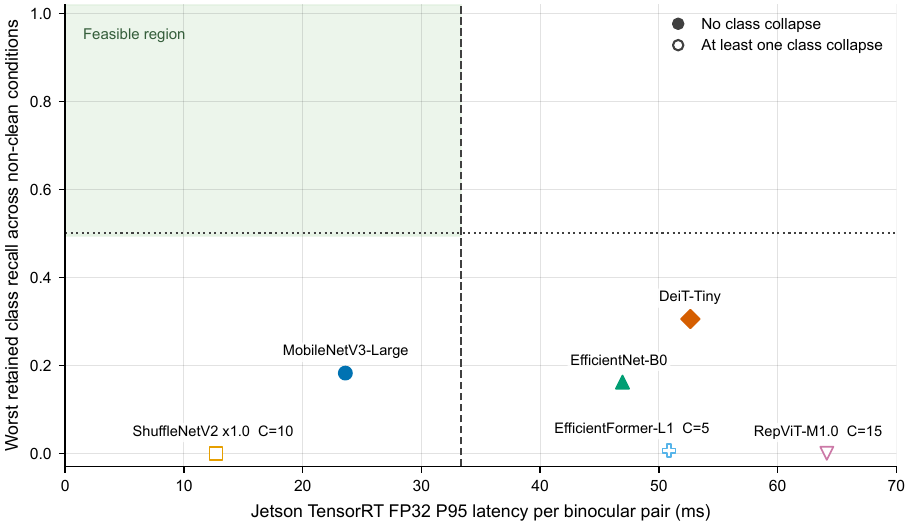}
    \caption{Primary operational constraint map. No model lies in the region satisfying both the 33.333-ms deadline and the 50\% worst-class recall floor.}
    \label{fig:primary_operational_constraint_map}
\end{figure}

Figure~\ref{fig:primary_operational_constraint_map} combines latency with the worst class recall retained over non-clean target-domain conditions. MobileNetV3-Large and ShuffleNetV2 satisfied the latency requirement but failed the safety-related screen; the other four failed both primary requirements. Consequently, $\mathcal{M}_{\mathrm{elig}}=\varnothing$, no primary Pareto set was constructed, and no threshold was relaxed after observing the outcome.

\subsection{RISE faithfulness and historical sensitivity}

\begin{figure}[!h]
    \centering
    \includegraphics[width=0.88\textwidth]{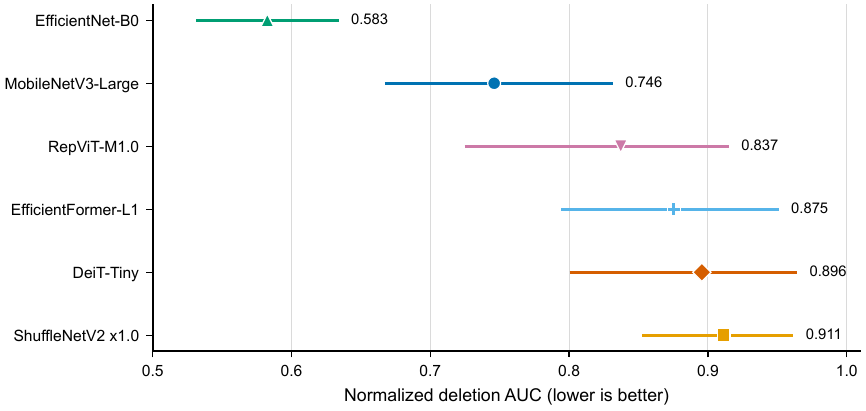}
    \caption{Subject-macro normalized deletion AUC with 95\% participant-bootstrap intervals. Lower values indicate faster target-score reduction when highly ranked regions are deleted.}
    \label{fig:rise_primary_faithfulness}
\end{figure}

Normalized deletion AUC ranged from 0.5826 for EfficientNet-B0 to 0.9113 for ShuffleNetV2 (Fig.~\ref{fig:rise_primary_faithfulness}); lower values are preferred. The subject-macro intervals were wide and overlapping for several models.

The quantitative comparison was complemented by two common-correct images selected without visual inspection. The two selected images form a different-participant pair that minimizes the combined distance between the six-model mean normalized deletion AUC and the corresponding class-specific medians (Fig.~\ref{fig:rise_qualitative_comparison}). An extended subject-stratified qualitative comparison is provided in Supplementary Figs.~S.7 and S.8. The open-eye gallery includes the four participants represented by eligible open-eye samples in the frozen XAI corpus; participant s0008 had no eligible open-eye sample.

\begin{figure}[!h]
    \centering
    \includegraphics[width=\textwidth]{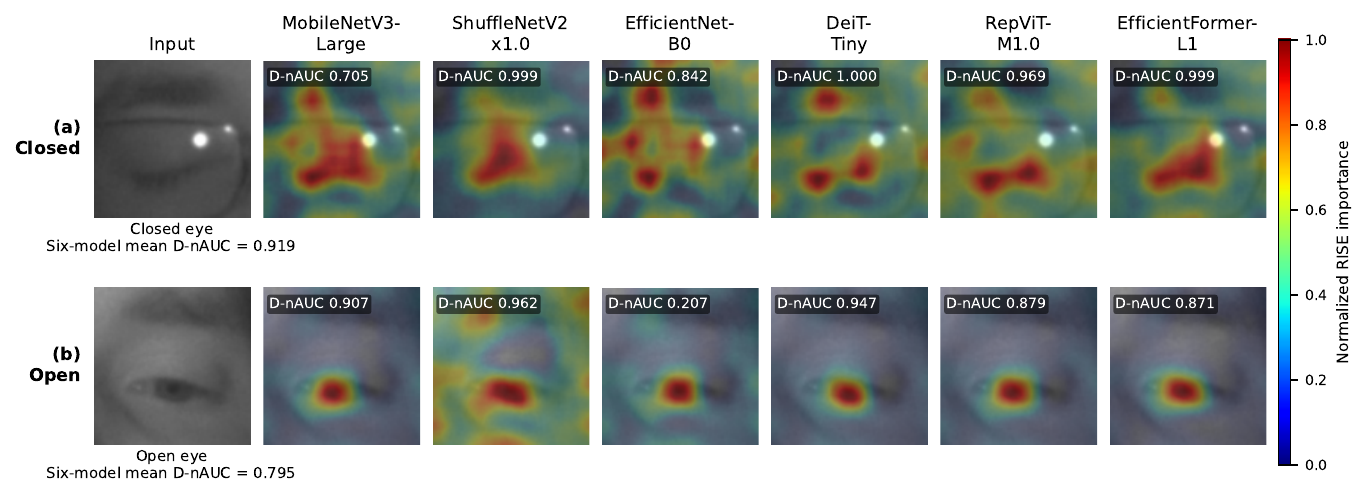}
    \caption{Qualitative RISE comparison for representative closed- and open-eye images correctly classified by all six models. D-nAUC denotes normalized deletion AUC; lower values are preferred. The examples are illustrative and should be interpreted with Fig.~\ref{fig:rise_primary_faithfulness}.}
    \label{fig:rise_qualitative_comparison}
\end{figure}

Insertion validity varied from 17.6\% to 88.6\%; only 49 images were valid for all six models, and 31 belonged to one participant. Its coverage profile is therefore reported in Supplementary Fig. S.2. Model-specific pixelwise Spearman and top-10\% intersection-over-union stability summaries are provided in Supplementary Figs.~S.3 and S.4. Stability across independent mask banks was moderate. Exact subject-level comparisons were exploratory because only five independent participants were available, making 0.0625 the minimum attainable two-sided exact $p$-value.

In the retrospective four-model HCS audit, ShuffleNetV2 ranked first under the Safety, Deployment, and Balanced profiles and occupied 0.7457 of the complete weight lattice. Nevertheless, all 24 strict orderings appeared in the preference space. Detailed scores and first-place shares are provided in Supplementary Table S.7; these shares are not probabilities of objective superiority and do not replace the current eligibility result. Supplementary Figs.~S.5 and S.6 visualize the first-place allocation and the complete lattice rank distributions, respectively.
\section{Discussion}
\label{sec:discussion}

\subsection{Rankings are dimension dependent}

The clean results illustrate why one in-domain metric cannot define a stable preference. MobileNetV3-Large led clean Macro-F1, whereas DeiT-Tiny led balanced accuracy and one directional error. Such differences are consistent with underspecification: models with similar held-out performance can implement different functions and fail differently after deployment \cite{damour2022underspecification}. Calibration and class-conditional errors therefore complement, rather than duplicate, aggregate discrimination.

Corruption and transfer changed the ordering again. DeiT-Tiny retained both classes and degraded progressively under Gaussian noise, whereas several other architectures developed class-specific collapse. Large-scale studies have often found vision transformers comparatively robust \cite{paul2022robustlearners,mao2022robustvit}, but the EfficientFormer-L1 and RepViT results do not support a universal transformer advantage. Robustness depended on the complete architecture and perturbation family. The weaker relation between clean and Gaussian-noise behavior also parallels findings that some shifts depart from the usual in-distribution/out-of-distribution accuracy association \cite{miller2021accuracyline}.

RT-BENE provided different evidence from controlled corruptions because it combined acquisition modality, illumination, head pose, prevalence, participant changes, and a change in the definition of the non-open target state. The Gray3 control further showed that transfer behavior was sensitive to color representation in an architecture-dependent manner: removing chromaticity changed both performance magnitude and model ordering. This ablation does not isolate sensing modality or reproduce near-infrared image formation, but it indicates that input representation contributes to the compound domain shift. Synthetic robustness is known to transfer incompletely to natural shifts \cite{taori2020naturalshift,koh2021wilds}, and backbone choice can alter domain generalization \cite{angarano2024backbones}. The strong DeiT-Tiny RGB zero-shot result is therefore evidence for architecture-dependent transfer within this protocol, not proof of a general transformer mechanism.

\subsection{Target-domain training changes the error allocation}

The largest target-domain gains occurred for models with weak zero-shot profiles. However, Macro-F1 improvement did not imply improvement in both error directions. ShuffleNetV2 recovered mainly by reducing open-to-blink errors while increasing blink-to-open errors; MobileNetV3-Large and EfficientNet-B0 showed the same trade-off. This is consistent with domain-generalization benchmarks showing that conclusions depend on the backbone and model-selection protocol \cite{gulrajani2021domainbed,zhou2023domain}. Even under matched frames and participant-safe evaluation, target-domain training remained architecture dependent.

Target-domain training should consequently not be treated as an automatic repair. It may recover substantial average performance while transferring the remaining error toward the less frequent class or preserving corruption vulnerabilities. Operational threshold design and target-domain training would need to be assessed jointly for a specific deployment.

\subsection{Efficiency must be measured on the target stack}

The Jetson results confirm that parameter count and desktop speed are imperfect proxies for embedded latency. Hardware-aware architectures were developed around this distinction: ShuffleNetV2 emphasized memory access \cite{ma2018shufflenetv2}, MnasNet incorporated measured mobile latency \cite{tan2019mnasnet}, and ProxylessNAS searched directly on the target hardware \cite{cai2019proxylessnas}. HCBF applies the same principle to benchmarking by timing compiled artifacts on the intended platform.

P95 latency also provides a stricter interpretation than the mean because real-time schedules depend on tail behavior. At the same time, the values reported here cover only the classifier engine, not image capture, eye localization, temporal fusion, or communication. They are therefore deployment evidence for the audited workload rather than end-to-end cabin latency.

\subsection{Faithfulness is conditional evidence}

RISE produced another ranking, with EfficientNet-B0 obtaining the lowest normalized deletion AUC. This does not establish universal explanation quality. Perturbation metrics depend on the removal baseline and can expose the model to unrealistic inputs \cite{gomez2022metrics, agarwal2021generativeremoval,nieradzik2025reliable}. Unequal insertion coverage further showed that a formally defined metric was not equally available across models. Comparing model-specific valid cohorts would confound faithfulness with sample selection.

The moderate stability across independent mask banks and the five-subject limit justify retaining RISE as diagnostic evidence rather than a gate or historical HCS component. The qualitative maps illustrate attribution patterns but do not establish causality, semantic correctness, or human usefulness.

\subsection{Ranking and eligibility answer different questions}

The historical HCS demonstrates the value and limitation of weighted scalarization. ShuffleNetV2 occupied most of the audited preference space, but all strict orderings appeared under some admissible weights. Weighted scores encode preferences, while Pareto methods preserve trade-offs \cite{emmerich2018multiobjective}; neither guarantees that an absolute requirement is met.

The non-compensatory gates answer a different question. Two models satisfied the latency requirement but failed the safety-related screen, whereas four failed both primary requirements. The empty eligible set is therefore not a failure to rank the portfolio; it indicates that no candidate satisfied all stated requirements. Restricting Pareto analysis to eligible models gives feasibility priority over preference and avoids presenting an inadmissible frontier.

More broadly, pattern-recognition benchmarks should report evidence in layers: predictive performance, robustness and transfer, target-hardware behavior, faithfulness, preference sensitivity, and mandatory eligibility. Directional errors, metric validity, and subject-level resampling may not improve the headline score, but they reduce the risk of overstating certainty. Under the evaluated conditions, HCBF identifies the specific improvements required by each architecture rather than forcing a universal winner.
\section{Limitations}
\label{sec:limitations}

The empirical task is binary eye-state recognition, not end-to-end drowsiness assessment. MRL provides isolated near-infrared crops and RT-BENE provides open/blink binocular frames; neither alone captures temporal closure, PERCLOS windows, gaze, head pose, vehicle dynamics, or physiological state. The source and target non-open labels are related but not identical: MRL Eye uses closed-eye annotations, whereas RT-BENE uses blink frames. Consequently, the external-domain results reflect both acquisition shift and a change in target-state semantics.

Independent subject support is also limited. The MRL test set contains five participants, so participant-bootstrap intervals are descriptive of those subjects rather than precise population estimates. This is especially restrictive for RISE, where the minimum attainable two-sided exact $p$-value was 0.0625. Training used fixed seeds and one common protocol; evaluation bootstrap does not include variability from repeated initialization, data ordering, or alternative hyperparameters.

RT-BENE is one compound external domain, not a comprehensive cross-dataset study. Its shift combines sensing, illumination, pose, prevalence, and participant changes. The corruption catalogue is likewise controlled but not exhaustive: compression, defocus, occlusion, reflections, localization error, combined corruptions, and sensor dropout were not evaluated. The grayscale control is a chroma ablation, not a physical simulation of near-infrared sensing.

Deployment evidence is specific to Jetson Nano, TensorRT FP32, and the audited power, clock, cooling, and batch configuration. It excludes energy and end-to-end pipeline latency, and it does not establish rankings on other hardware, precisions, or runtimes. Numerical parity was established on frozen validation inputs and should not be interpreted as a full-test-set equivalence evaluation. RISE is conditional on one explainer, baseline, mask distribution, and target; insertion had unequal coverage and no human evaluation of explanation usefulness was conducted.

Finally, the HCS weights and operational gates encode explicit choices. The HCS remains a retrospective four-model audit, and its first-place share is not a probability of superiority. The latency and classwise-recall thresholds are prespecified benchmarking requirements for the primary profile. The safety-related screen is comparative and should not be interpreted as automotive safety certification, a regulatory limit, or evidence that a model satisfying the screen would be sufficient for deployment. The six architectures are representative rather than exhaustive, so the results support the methodological claim that ranking and eligibility can change across dimensions, not a permanent ordering of model families.
\section{Conclusions}
\label{sec:conclusions}

HCBF was developed to determine whether model preference remains stable when vision-based eye-state recognition is evaluated beyond clean aggregate performance. The framework preserves distinct evidence from prediction, directional robustness, cross-domain behavior, target-hardware execution, and black-box faithfulness, while separating descriptive preference from mandatory eligibility.

The six architectures changed order across dimensions. MobileNetV3-Large led clean Macro-F1, DeiT-Tiny led zero-shot transfer and retained the most progressive Gaussian-noise response, ShuffleNetV2 gained most from target-domain training and had the lowest embedded latency, and EfficientNet-B0 had the lowest normalized deletion AUC. None of these advantages established universal superiority.

Most importantly, no model satisfied both the latency and safety-related requirements. The eligible set was retained as empty without relaxing the predefined gates or constructing a Pareto frontier over ineligible models. The historical HCS remained useful for preference sensitivity, but not for deployment readiness.

Future work should include repeated training seeds, additional external datasets and modalities, naturally observed and combined corruptions, temporal or multimodal models, newer embedded platforms, reduced precision, energy and end-to-end timing, and broader explanation audits. The central conclusion is methodological: deployment-aware model selection should preserve directional failure and uncertainty and should allow no model to be selected when the available portfolio does not meet the stated requirements.


\section*{Funding}

This research did not receive any specific grant from funding agencies in the
public, commercial, or not-for-profit sectors.

\section*{Declaration of competing interest}

The author declares that he has no known competing financial interests or
personal relationships that could have appeared to influence the work reported
in this article.

\section*{Data availability}

The MRL Eye and RT-BENE datasets are publicly available from their original
providers. Processed tables, frozen evaluation protocols, manifests,
supplementary results, and versioned analysis resources supporting this study
are available in the companion GitHub repository:
\url{https://github.com/rubendflorezzela/hcbf-driver-monitoring}.
The frozen software and analysis package corresponding to version v2.0.0 has
been archived on Zenodo:
\url{https://doi.org/10.5281/zenodo.21908706}.










\bibliographystyle{elsarticle-num}
\bibliography{references}

@article{qu2024driver,
  author  = {Fangming Qu and Nolan Dang and Borko Furht and Mehrdad Nojoumian},
  title   = {Comprehensive Study of Driver Behavior Monitoring Systems Using Computer Vision and Machine Learning Techniques},
  journal = {Journal of Big Data},
  volume  = {11},
  number  = {1},
  pages   = {32},
  year    = {2024},
  doi     = {10.1186/s40537-024-00890-0}
}

@article{albadawi2022review,
  author  = {Yaman Albadawi and Maen Takruri and Mohammed Awad},
  title   = {A Review of Recent Developments in Driver Drowsiness Detection Systems},
  journal = {Sensors},
  volume  = {22},
  number  = {5},
  pages   = {2069},
  year    = {2022},
  doi     = {10.3390/s22052069}
}

@article{kolus2024eye,
  author  = {Ahmet Kolus},
  title   = {A Systematic Review on Driver Drowsiness Detection Using Eye Activity Measures},
  journal = {IEEE Access},
  volume  = {12},
  pages   = {97969--97993},
  year    = {2024},
  doi     = {10.1109/ACCESS.2024.3424654}
}

@article{abe2023perclos,
  author  = {Takashi Abe},
  title   = {{PERCLOS}-Based Technologies for Detecting Drowsiness: Current Evidence and Future Directions},
  journal = {Sleep Advances},
  volume  = {4},
  number  = {1},
  pages   = {zpad006},
  year    = {2023},
  doi     = {10.1093/sleepadvances/zpad006}
}

@inproceedings{cortacero2019rtbene,
  author    = {Kevin Cortacero and Tobias Fischer and Yiannis Demiris},
  title     = {{RT-BENE}: A Dataset and Baselines for Real-Time Blink Estimation in Natural Environments},
  booktitle = {Proceedings of the IEEE/CVF International Conference on Computer Vision Workshops},
  year      = {2019},
  doi       = {10.1109/ICCVW.2019.00147}
}

@article{damour2022underspecification,
  author  = {Alexander D'Amour and Katherine Heller and Dan Moldovan and Ben Adlam and Babak Alipanahi and Alex Beutel and Christina Chen and Jonathan Deaton and Jacob Eisenstein and Matthew D. Hoffman and Farhad Hormozdiari and Neil Houlsby and Shaobo Hou and Ghassen Jerfel and Alan Karthikesalingam and Mario Lucic and Yian Ma and Cory McLean and Diana Mincu and Akinori Mitani and Andrea Montanari and Zachary Nado and Vivek Natarajan and Christopher Nielson and Thomas F. Osborne and Rajiv Raman and Kim Ramasamy and Rory Sayres and Jessica Schrouff and Martin Seneviratne and Shannon Sequeira and Harini Suresh and Victor Veitch and Max Vladymyrov and Xuezhi Wang and Kellie Webster and Steve Yadlowsky and Taedong Yun and Xiaohua Zhai and D. Sculley},
  title   = {Underspecification Presents Challenges for Credibility in Modern Machine Learning},
  journal = {Journal of Machine Learning Research},
  volume  = {23},
  number  = {226},
  pages   = {1--61},
  year    = {2022}
}

@inproceedings{guo2017calibration,
  author    = {Chuan Guo and Geoff Pleiss and Yu Sun and Kilian Q. Weinberger},
  title     = {On Calibration of Modern Neural Networks},
  booktitle = {Proceedings of the 34th International Conference on Machine Learning},
  series    = {Proceedings of Machine Learning Research},
  volume    = {70},
  pages     = {1321--1330},
  year      = {2017}
}

@inproceedings{hendrycks2019corruptions,
  author    = {Dan Hendrycks and Thomas Dietterich},
  title     = {Benchmarking Neural Network Robustness to Common Corruptions and Perturbations},
  booktitle = {International Conference on Learning Representations},
  year      = {2019}
}

@inproceedings{taori2020naturalshift,
  author    = {Rohan Taori and Achal Dave and Vaishaal Shankar and Nicholas Carlini and Benjamin Recht and Ludwig Schmidt},
  title     = {Measuring Robustness to Natural Distribution Shifts in Image Classification},
  booktitle = {Advances in Neural Information Processing Systems},
  volume    = {33},
  pages     = {18583--18599},
  year      = {2020}
}

@article{zhou2023domain,
  author  = {Kaiyang Zhou and Ziwei Liu and Yu Qiao and Tao Xiang and Chen Change Loy},
  title   = {Domain Generalization: A Survey},
  journal = {IEEE Transactions on Pattern Analysis and Machine Intelligence},
  volume  = {45},
  number  = {4},
  pages   = {4396--4415},
  year    = {2023},
  doi     = {10.1109/TPAMI.2022.3195549}
}

@article{wang2023domain,
  author  = {Jindong Wang and Cuiling Lan and Chang Liu and Yidong Ouyang and Tao Qin and Wang Lu and Yiqiang Chen and Wenjun Zeng and Philip S. Yu},
  title   = {Generalizing to Unseen Domains: A Survey on Domain Generalization},
  journal = {IEEE Transactions on Knowledge and Data Engineering},
  volume  = {35},
  number  = {8},
  pages   = {8052--8072},
  year    = {2023},
  doi     = {10.1109/TKDE.2022.3178128}
}

@incollection{ma2018shufflenetv2,
  author    = {Ningning Ma and Xiangyu Zhang and Hai-Tao Zheng and Jian Sun},
  title     = {{ShuffleNet V2}: Practical Guidelines for Efficient {CNN} Architecture Design},
  booktitle = {Computer Vision -- ECCV 2018},
  pages     = {122--138},
  publisher = {Springer},
  year      = {2018},
  doi       = {10.1007/978-3-030-01264-9_8}
}

@inproceedings{howard2019mobilenetv3,
  author    = {Andrew Howard and Mark Sandler and Grace Chu and Liang-Chieh Chen and Bo Chen and Mingxing Tan and Weijun Wang and Yukun Zhu and Ruoming Pang and Vijay Vasudevan and Quoc V. Le and Hartwig Adam},
  title     = {Searching for {MobileNetV3}},
  booktitle = {Proceedings of the IEEE/CVF International Conference on Computer Vision},
  pages     = {1314--1324},
  year      = {2019}
}

@inproceedings{adebayo2018sanity,
  author    = {Julius Adebayo and Justin Gilmer and Michael Muelly and Ian Goodfellow and Moritz Hardt and Been Kim},
  title     = {Sanity Checks for Saliency Maps},
  booktitle = {Advances in Neural Information Processing Systems},
  volume    = {31},
  pages     = {9525--9536},
  year      = {2018}
}

@inproceedings{hooker2019benchmark,
  author    = {Sara Hooker and Dumitru Erhan and Pieter-Jan Kindermans and Been Kim},
  title     = {A Benchmark for Interpretability Methods in Deep Neural Networks},
  booktitle = {Advances in Neural Information Processing Systems},
  volume    = {32},
  pages     = {9737--9748},
  articleno = {873},
  year      = {2019}
}

@inproceedings{petsiuk2018rise,
  author    = {Vitali Petsiuk and Abir Das and Kate Saenko},
  title     = {{RISE}: Randomized Input Sampling for Explanation of Black-Box Models},
  booktitle = {British Machine Vision Conference},
  year      = {2018}
}

@article{song2014eyes,
  author  = {Feng Song and Xiaoyang Tan and Xue Liu and Songcan Chen},
  title   = {Eyes Closeness Detection from Still Images with Multi-Scale Histograms of Principal Oriented Gradients},
  journal = {Pattern Recognition},
  volume  = {47},
  number  = {9},
  pages   = {2825--2838},
  year    = {2014},
  doi     = {10.1016/j.patcog.2014.03.024}
}

@incollection{fusek2018pupil,
  author    = {Radovan Fusek},
  title     = {Pupil Localization Using Geodesic Distance},
  booktitle = {Advances in Visual Computing},
  series    = {Lecture Notes in Computer Science},
  volume    = {11241},
  pages     = {433--444},
  publisher = {Springer},
  address   = {Cham},
  year      = {2018},
  doi       = {10.1007/978-3-030-03801-4_38}
}

@incollection{ortega2021dmd,
  author    = {Juan Diego Ortega and Neslihan Kose and Paola Cañas and Min-An Chao and Alexander Unnervik and Marcos Nieto and Oihana Otaegui and Luis Salgado},
  title     = {{DMD}: A Large-Scale Multi-Modal Driver Monitoring Dataset for Attention and Alertness Analysis},
  booktitle = {Computer Vision -- ECCV 2020 Workshops},
  series    = {Lecture Notes in Computer Science},
  pages     = {387--405},
  publisher = {Springer},
  address   = {Cham},
  year      = {2020},
  doi       = {10.1007/978-3-030-66823-5_23}
}

@inproceedings{gulrajani2021domainbed,
  author    = {Ishaan Gulrajani and David Lopez-Paz},
  title     = {In Search of Lost Domain Generalization},
  booktitle = {International Conference on Learning Representations},
  year      = {2021}
}

@inproceedings{koh2021wilds,
  author    = {Pang Wei Koh and Shiori Sagawa and Henrik Marklund and Sang Michael Xie and Marvin Zhang and Akshay Balsubramani and Weihua Hu and Michihiro Yasunaga and Richard Lanas Phillips and Irena Gao and Tony Lee and Etienne David and Ian Stavness and Wei Guo and Berton Earnshaw and Imran Haque and Sara M. Beery and Jure Leskovec and Anshul Kundaje and Emma Pierson and Sergey Levine and Chelsea Finn and Percy Liang},
  title     = {{WILDS}: A Benchmark of in-the-Wild Distribution Shifts},
  booktitle = {Proceedings of the 38th International Conference on Machine Learning},
  series    = {Proceedings of Machine Learning Research},
  volume    = {139},
  pages     = {5637--5664},
  year      = {2021},
  publisher = {PMLR}
}

@article{angarano2024backbones,
  author  = {Simone Angarano and Mauro Martini and Francesco Salvetti and Vittorio Mazzia and Marcello Chiaberge},
  title   = {Back-to-Bones: Rediscovering the Role of Backbones in Domain Generalization},
  journal = {Pattern Recognition},
  volume  = {156},
  pages   = {110762},
  year    = {2024},
  doi     = {10.1016/j.patcog.2024.110762}
}

@inproceedings{tan2019efficientnet,
  author    = {Mingxing Tan and Quoc V. Le},
  title     = {{EfficientNet}: Rethinking Model Scaling for Convolutional Neural Networks},
  booktitle = {Proceedings of the 36th International Conference on Machine Learning},
  series    = {Proceedings of Machine Learning Research},
  volume    = {97},
  pages     = {6105--6114},
  year      = {2019},
  publisher = {PMLR}
}

@inproceedings{touvron2021deit,
  author    = {Hugo Touvron and Matthieu Cord and Matthijs Douze and Francisco Massa and Alexandre Sablayrolles and Hervé Jégou},
  title     = {Training Data-Efficient Image Transformers and Distillation through Attention},
  booktitle = {Proceedings of the 38th International Conference on Machine Learning},
  series    = {Proceedings of Machine Learning Research},
  volume    = {139},
  pages     = {10347--10357},
  year      = {2021},
  publisher = {PMLR}
}

@inproceedings{li2022efficientformer,
  author    = {Yanyu Li and Geng Yuan and Yang Wen and Ju Hu and Georgios Evangelidis and Sergey Tulyakov and Yanzhi Wang and Jian Ren},
  title     = {{EfficientFormer}: Vision Transformers at {MobileNet} Speed},
  booktitle = {Advances in Neural Information Processing Systems},
  volume    = {35},
  pages     = {12934--12949},
  year      = {2022}
}

@inproceedings{wang2024repvit,
  author    = {Ao Wang and Hui Chen and Zijia Lin and Jungong Han and Guiguang Ding},
  title     = {{RepViT}: Revisiting Mobile {CNN} from {ViT} Perspective},
  booktitle = {Proceedings of the IEEE/CVF Conference on Computer Vision and Pattern Recognition},
  pages     = {15909--15920},
  year      = {2024},
  doi       = {10.1109/CVPR52733.2024.01506}
}

@inproceedings{selvaraju2017gradcam,
  author    = {Ramprasaath R. Selvaraju and Michael Cogswell and Abhishek Das and Ramakrishna Vedantam and Devi Parikh and Dhruv Batra},
  title     = {{Grad-CAM}: Visual Explanations from Deep Networks via Gradient-Based Localization},
  booktitle = {Proceedings of the IEEE International Conference on Computer Vision},
  pages     = {618--626},
  year      = {2017},
  doi       = {10.1109/ICCV.2017.74}
}

@incollection{gomez2022metrics,
  author    = {Tristan Gomez and Thomas Fréour and Harold Mouchère},
  title     = {Metrics for Saliency Map Evaluation of Deep Learning Explanation Methods},
  booktitle = {Pattern Recognition and Artificial Intelligence},
  series    = {Lecture Notes in Computer Science},
  volume    = {13363},
  pages     = {84--95},
  publisher = {Springer},
  address   = {Cham},
  year      = {2022},
  doi       = {10.1007/978-3-031-09037-0_8}
}

@inproceedings{mitchell2019modelcards,
  author    = {Margaret Mitchell and Simone Wu and Andrew Zaldivar and Parker Barnes and Lucy Vasserman and Ben Hutchinson and Elena Spitzer and Inioluwa Deborah Raji and Timnit Gebru},
  title     = {Model Cards for Model Reporting},
  booktitle = {Proceedings of the Conference on Fairness, Accountability, and Transparency},
  pages     = {220--229},
  year      = {2019},
  publisher = {ACM},
  doi       = {10.1145/3287560.3287596}
}

@inproceedings{paul2022robustlearners,
  author    = {Sayak Paul and Pin-Yu Chen},
  title     = {Vision Transformers Are Robust Learners},
  booktitle = {Proceedings of the AAAI Conference on Artificial Intelligence},
  volume    = {36},
  number    = {2},
  pages     = {2071--2081},
  year      = {2022},
  doi       = {10.1609/aaai.v36i2.20103}
}

@inproceedings{mao2022robustvit,
  author    = {Xiaofeng Mao and Gege Qi and Yuefeng Chen and Xiaodan Li and
               Ranjie Duan and Shaokai Ye and Yuan He and Hui Xue},
  title     = {Towards Robust Vision Transformer},
  booktitle = {Proceedings of the IEEE/CVF Conference on Computer Vision and
               Pattern Recognition},
  pages     = {12032--12041},
  year      = {2022}
}

@inproceedings{miller2021accuracyline,
  author    = {John P. Miller and Rohan Taori and Aditi Raghunathan and
               Shiori Sagawa and Pang Wei Koh and Vaishaal Shankar and
               Percy Liang and Yair Carmon and Ludwig Schmidt},
  title     = {Accuracy on the Line: On the Strong Correlation Between
               Out-of-Distribution and In-Distribution Generalization},
  booktitle = {Proceedings of the 38th International Conference on Machine
               Learning},
  series    = {Proceedings of Machine Learning Research},
  volume    = {139},
  pages     = {7721--7735},
  publisher = {PMLR},
  year      = {2021}
}

@inproceedings{tan2019mnasnet,
  author    = {Mingxing Tan and Bo Chen and Ruoming Pang and Vijay Vasudevan and
               Mark Sandler and Andrew Howard and Quoc V. Le},
  title     = {{MnasNet}: Platform-Aware Neural Architecture Search for Mobile},
  booktitle = {Proceedings of the IEEE/CVF Conference on Computer Vision and
               Pattern Recognition},
  pages     = {2815--2823},
  year      = {2019}
}

@inproceedings{cai2019proxylessnas,
  author    = {Han Cai and Ligeng Zhu and Song Han},
  title     = {{ProxylessNAS}: Direct Neural Architecture Search on Target Task
               and Hardware},
  booktitle = {International Conference on Learning Representations},
  year      = {2019}
}

@incollection{agarwal2021generativeremoval,
  author    = {Chirag Agarwal and Anh Nguyen},
  title     = {Explaining Image Classifiers by Removing Input Features Using
               Generative Models},
  booktitle = {Computer Vision -- ACCV 2020},
  series    = {Lecture Notes in Computer Science},
  volume    = {12627},
  pages     = {101--118},
  publisher = {Springer},
  address   = {Cham},
  year      = {2021},
  doi       = {10.1007/978-3-030-69544-6_7}
}

@article{nieradzik2025reliable,
  author  = {Lars Nieradzik and Henrike Stephani and Janis Keuper},
  title   = {Reliable Evaluation of Attribution Maps in {CNNs}: A
             Perturbation-Based Approach},
  journal = {International Journal of Computer Vision},
  volume  = {133},
  pages   = {2392--2409},
  year    = {2025},
  doi     = {10.1007/s11263-024-02282-6}
}

@article{emmerich2018multiobjective,
  author  = {Michael T. M. Emmerich and Andr{\'e} H. Deutz},
  title   = {A Tutorial on Multiobjective Optimization: Fundamentals and
             Evolutionary Methods},
  journal = {Natural Computing},
  volume  = {17},
  number  = {3},
  pages   = {585--609},
  year    = {2018},
  doi     = {10.1007/s11047-018-9685-y}
}

\end{document}


\maketitle

\noindent
This supplement provides the extended definitions, complete tabular results,
and secondary diagnostic figures referenced in the main manuscript. It does not redefine the frozen binary decision rule, operational gates, or
historical HCS scope.

\tableofcontents
\clearpage

\section{Extended metric definitions and aggregation}
\label{sec:supp_metrics}

\subsection{Prediction semantics and confusion-derived metrics}

For image or binocular frame $i$, the frozen binary decision is defined in
Eq.~\eqref{eq:supp_decision}.
\begin{equation}
\widehat y_{m,i}=\mathbf{1}[p_{m,i}\geq 0.5],
\label{eq:supp_decision}
\end{equation}
where $p_{m,i}$ is the open-eye probability, $y_i=1$ denotes open, and
$y_i=0$ denotes the non-open class. The non-open class is closed on MRL Eye
and blink on RT-BENE. For RT-BENE, $p_{m,i}$ is the arithmetic mean of the
left- and right-eye open probabilities.

Let $\mathrm{TP}_c$, $\mathrm{FP}_c$, and $\mathrm{FN}_c$ denote the
one-versus-rest counts for class $c\in\{0,1\}$. Class precision, recall, and F1 are defined in Eq.~\eqref{eq:supp_class_metrics}.
\begin{align}
P_c&=\frac{\mathrm{TP}_c}{\mathrm{TP}_c+\mathrm{FP}_c},&
R_c&=\frac{\mathrm{TP}_c}{\mathrm{TP}_c+\mathrm{FN}_c},&
F1_c&=\frac{2P_cR_c}{P_c+R_c}.
\label{eq:supp_class_metrics}
\end{align}
Macro-F1 and balanced accuracy are defined in Eq.~\eqref{eq:supp_macro_metrics}.
\begin{align}
F1^{\mathrm{macro}}&=\frac{F1_0+F1_1}{2},&
\mathrm{BAcc}&=\frac{R_0+R_1}{2}.
\label{eq:supp_macro_metrics}
\end{align}
When a resampled participant set lacks support for a required class, the
corresponding class-dependent statistic is left undefined rather than
imputed.

The two directional error rates retained throughout the evaluation are defined in Eq.~\eqref{eq:supp_directional_errors}.
\begin{align}
e_{0\rightarrow1}(m)&=
\frac{\sum_i\mathbf{1}[y_i=0,\widehat y_{m,i}=1]}
{\sum_i\mathbf{1}[y_i=0]},&
e_{1\rightarrow0}(m)&=
\frac{\sum_i\mathbf{1}[y_i=1,\widehat y_{m,i}=0]}
{\sum_i\mathbf{1}[y_i=1]}.
\label{eq:supp_directional_errors}
\end{align}
They are closed-to-open and open-to-closed on MRL, and blink-to-open and
open-to-blink on RT-BENE.

\subsection{Probability quality}

The Brier score is defined in Eq.~\eqref{eq:supp_brier}.
\begin{equation}
\operatorname{BS}_m=\frac{1}{N}\sum_{i=1}^{N}(p_{m,i}-y_i)^2.
\label{eq:supp_brier}
\end{equation}
For ECE, the predicted-class confidence $c_{m,i}=\max(p_{m,i},1-p_{m,i})$ is partitioned into 15 fixed-width bins, and the resulting measure is defined in Eq.~\eqref{eq:supp_ece}.
\begin{equation}
\operatorname{ECE}_m=
\sum_{b=1}^{15}\frac{|\mathcal I_b|}{N}
\left|\operatorname{acc}(\mathcal I_b)-
\operatorname{conf}(\mathcal I_b)\right|.
\label{eq:supp_ece}
\end{equation}
No post-training calibration was applied.

\subsection{Corruption and cross-domain effects}

For metric $q$ and corruption condition $c$, the matched effect relative to
the clean anchor is defined in Eq.~\eqref{eq:supp_corruption_effect}.
\begin{equation}
\Delta_{m,c}^{(q)}=q_{m,c}-q_{m,\mathrm{clean}}.
\label{eq:supp_corruption_effect}
\end{equation}
Gaussian-noise realizations were pooled within participant and severity.
For family $g$ with condition set $\mathcal{C}_g$, the family mean and
worst-family Macro-F1 are defined in Eq.~\eqref{eq:supp_family_aggregation}.
\begin{align}
\overline F_{m,g}&=
\frac{1}{|\mathcal C_g|}\sum_{c\in\mathcal C_g}
F^{\mathrm{macro}}_{m,c},&
F_m^{\mathrm{worst}}&=\min_g\overline F_{m,g}.
\label{eq:supp_family_aggregation}
\end{align}
The worst directional error is defined in Eq.~\eqref{eq:supp_worst_direction}.
\begin{equation}
e_m^{\mathrm{worst}}=
\max_c\max\{e_{0\rightarrow1}(m;c),e_{1\rightarrow0}(m;c)\}.
\label{eq:supp_worst_direction}
\end{equation}

The frozen collapse indicator flags an all-one-class decision or a predicted
class fraction of at least 0.95. Collapse is retained as a separate
diagnostic even when an aggregate metric remains finite.

For the matched target-domain comparison, the effect is defined in
Eq.~\eqref{eq:supp_adaptation_effect}.
\begin{equation}
A_{m,c}^{(q)}=
q_{m,c}^{(\mathrm{OOF})}-q_{m,c}^{(0)},
\label{eq:supp_adaptation_effect}
\end{equation}
where both values use the same frames and participants.

\subsection{Participant-cluster uncertainty}

Participants are the independent resampling units. A bootstrap replicate,
defined in Eq.~\eqref{eq:supp_cluster_bootstrap}, is formed by sampling
participant identifiers with replacement and concatenating their complete
contributions.
\begin{equation}
\widehat\theta^{*(b)}=
T\!\left(
\operatorname{concat}
[\mathcal D_{I_1^{(b)}},\ldots,\mathcal D_{I_S^{(b)}}]
\right).
\label{eq:supp_cluster_bootstrap}
\end{equation}
Shared participant weights are used for paired model, condition, and regime
contrasts. Reported intervals are percentile 95\% intervals from 10,000
replicates.

\section{Historical HCS formulation and sensitivity}
\label{sec:supp_hcs}

The HCS audit is retrospective and restricted to MobileNetV3-Large,
ShuffleNetV2 x1.0, EfficientNet-B0, and DeiT-Tiny. Its dimensions are
historical accuracy $\alpha_m$, legacy explainability $\epsilon_m$,
efficiency $\eta_m$, and robustness $\rho_m$. RepViT-M1.0,
EfficientFormer-L1, and the current RISE evidence are not inserted
retrospectively.

For historical quantity $z_m$, min--max normalization over the original
four-model set is defined in Eq.~\eqref{eq:supp_minmax}.
\begin{equation}
\mathcal N(z_m)=
\frac{z_m-\min_j z_j}{\max_j z_j-\min_j z_j}.
\label{eq:supp_minmax}
\end{equation}
If the range is zero, the normalized term is set to zero. With parameter count $P_m$, FLOPs $F_m$, and reference latency $L_m$,
historical efficiency is defined in Eq.~\eqref{eq:supp_efficiency}.
\begin{equation}
\eta_m=
1-\frac{\mathcal N(P_m)+\mathcal N(F_m)+\mathcal N(L_m)}{3}.
\label{eq:supp_efficiency}
\end{equation}
For nonnegative weights summing to one, the historical HCS is defined in
Eq.~\eqref{eq:supp_hcs}.
\begin{equation}
\operatorname{HCS}_m(\mathbf w)=
w_\alpha\alpha_m+w_\epsilon\epsilon_m+
w_\eta\eta_m+w_\rho\rho_m.
\label{eq:supp_hcs}
\end{equation}
The original Safety, Deployment, and Balanced profiles are specified in
Eq.~\eqref{eq:supp_hcs_profiles}.
\begin{align}
\mathbf w^{\mathrm{Safety}}&=(0.30,0.20,0.20,0.30),\nonumber\\
\mathbf w^{\mathrm{Deployment}}&=(0.25,0.20,0.35,0.20),\nonumber\\
\mathbf w^{\mathrm{Balanced}}&=(0.25,0.25,0.25,0.25).
\label{eq:supp_hcs_profiles}
\end{align}

The complete nonnegative simplex was enumerated in increments of 0.01,
yielding 176,851 weight vectors. A deterministic sample of 250,000 vectors
from $\operatorname{Dirichlet}(1,1,1,1)$ independently checked the continuous
simplex. Tied first positions are allocated fractionally according to
Eq.~\eqref{eq:supp_first_share}.
\begin{equation}
\pi_m=
\frac{1}{|\mathcal W|}
\sum_{\mathbf w\in\mathcal W}
\frac{\mathbf{1}[m\in\arg\max_j\operatorname{HCS}_j(\mathbf w)]}
{|\arg\max_j\operatorname{HCS}_j(\mathbf w)|}.
\label{eq:supp_first_share}
\end{equation}
The value $\pi_m$ measures coverage of the audited preference space; it is not
a probability that model $m$ is objectively superior.

\section{Supplementary experimental evidence}
\label{sec:supp_evidence}

\subsection{Models and complete predictive metrics}

The frozen model and checkpoint metadata are reported in
Table~\ref{tab:supp_model_metadata}. All six checkpoints passed the finite
output and strict-loading checks used before evaluation.

\begin{landscape}

\begin{table}[!h]
\centering
\caption{Architecture and frozen-checkpoint metadata. The SHA-256 column shows
the first 12 hexadecimal characters; complete hashes are provided in the
machine-readable file accompanying this supplement.}
\label{tab:supp_model_metadata}
\small
\setlength{\tabcolsep}{4pt}
\begin{tabularx}{\textwidth}{@{}p{3.1cm}Xrcccc@{}}
\toprule
Architecture & Design family & Parameters & Best epoch & Stop epoch &
Best val. Macro-F1 & SHA-256 prefix \\
\midrule
MobileNetV3-Large & Mobile CNN & 4,204,594 & 11 & 16 & 0.973484 & \texttt{197170060127} \\
ShuffleNetV2 x1.0 & Mobile CNN & 1,255,654 & 11 & 16 & 0.976821 & \texttt{6051f8605112} \\
EfficientNet-B0 & Scaled CNN & 4,010,110 & 6 & 11 & 0.972425 & \texttt{213756c9c372} \\
DeiT-Tiny & Compact transformer & 5,524,802 & 13 & 18 & 0.972527 & \texttt{07de216dd9b3} \\
RepViT-M1.0 & Modern mobile CNN & 6,404,798 & 17 & 22 & 0.972231 & \texttt{bbab98455f50} \\
EfficientFormer-L1 & Mobile transformer & 11,393,724 & 14 & 19 & 0.975329 & \texttt{6f250a6019c1} \\
\bottomrule
\end{tabularx}
\end{table}

Table~\ref{tab:supp_complete_predictive_metrics} expands the predictive
summary in the main manuscript by reporting calibration, class-specific
precision, recall, F1, and both directional errors.
\begin{table}[!h]
\centering
\caption{Complete predictive metrics for the clean MRL test partition and the
zero-shot RT-BENE binocular-frame cohort. Non-open denotes closed on MRL and
blink on RT-BENE.}
\label{tab:supp_complete_predictive_metrics}
\tiny
\setlength{\tabcolsep}{2.2pt}

\textbf{Panel A: clean MRL test}\\[2mm]
\resizebox{\linewidth}{!}{%
\begin{tabular}{lrrrrrrrrrrrrrr}
\toprule
Model & Acc. & Macro-F1 & BAcc & AUROC & Brier & ECE &
P$_0$ & R$_0$ & F1$_0$ & P$_1$ & R$_1$ & F1$_1$ &
$e_{0\rightarrow1}$ & $e_{1\rightarrow0}$ \\
\midrule
MobileNetV3-Large & 0.9899 & 0.9794 & 0.9777 & 0.9989 & 0.0079 & 0.0022 & 0.9934 & 0.9948 & 0.9941 & 0.9686 & 0.9607 & 0.9646 & 0.0052 & 0.0393 \\
ShuffleNetV2 x1.0 & 0.9793 & 0.9566 & 0.9417 & 0.9976 & 0.0153 & 0.0115 & 0.9816 & 0.9945 & 0.9880 & 0.9646 & 0.8888 & 0.9252 & 0.0055 & 0.1112 \\
EfficientNet-B0 & 0.9819 & 0.9625 & 0.9533 & 0.9954 & 0.0141 & 0.0035 & 0.9855 & 0.9934 & 0.9895 & 0.9588 & 0.9133 & 0.9355 & 0.0066 & 0.0867 \\
DeiT-Tiny & 0.9875 & 0.9752 & 0.9844 & 0.9983 & 0.0101 & 0.0018 & 0.9966 & 0.9888 & 0.9927 & 0.9363 & 0.9800 & 0.9576 & 0.0112 & 0.0200 \\
RepViT-M1.0 & 0.9822 & 0.9631 & 0.9538 & 0.9968 & 0.0138 & 0.0055 & 0.9857 & 0.9936 & 0.9896 & 0.9603 & 0.9140 & 0.9366 & 0.0064 & 0.0860 \\
EfficientFormer-L1 & 0.9890 & 0.9780 & 0.9834 & 0.9986 & 0.0085 & 0.0040 & 0.9959 & 0.9913 & 0.9936 & 0.9495 & 0.9755 & 0.9623 & 0.0087 & 0.0245 \\
\bottomrule
\end{tabular}
}

\vspace{5mm}
\textbf{Panel B: zero-shot RT-BENE}\\[2mm]
\resizebox{\linewidth}{!}{%
\begin{tabular}{lrrrrrrrrrrrrrr}
\toprule
Model & Acc. & Macro-F1 & BAcc & AUROC & Brier & ECE &
P$_0$ & R$_0$ & F1$_0$ & P$_1$ & R$_1$ & F1$_1$ &
$e_{0\rightarrow1}$ & $e_{1\rightarrow0}$ \\
\midrule
MobileNetV3-Large & 0.8219 & 0.6034 & 0.8774 & 0.9063 & 0.1508 & 0.1003 & 0.1850 & 0.9381 & 0.3090 & 0.9967 & 0.8168 & 0.8978 & 0.0619 & 0.1832 \\
ShuffleNetV2 x1.0 & 0.2216 & 0.2066 & 0.5885 & 0.6713 & 0.6183 & 0.6155 & 0.0512 & 0.9895 & 0.0974 & 0.9975 & 0.1876 & 0.3158 & 0.0105 & 0.8124 \\
EfficientNet-B0 & 0.7712 & 0.5551 & 0.8205 & 0.8737 & 0.1516 & 0.0295 & 0.1424 & 0.8745 & 0.2450 & 0.9928 & 0.7666 & 0.8651 & 0.1255 & 0.2334 \\
DeiT-Tiny & 0.9456 & 0.7771 & 0.9228 & 0.9775 & 0.0404 & 0.0120 & 0.4320 & 0.8980 & 0.5834 & 0.9952 & 0.9477 & 0.9709 & 0.1020 & 0.0523 \\
RepViT-M1.0 & 0.8699 & 0.6387 & 0.8480 & 0.9440 & 0.0872 & 0.0542 & 0.2219 & 0.8240 & 0.3497 & 0.9911 & 0.8719 & 0.9277 & 0.1760 & 0.1281 \\
EfficientFormer-L1 & 0.8785 & 0.6314 & 0.7949 & 0.9018 & 0.0935 & 0.0383 & 0.2152 & 0.7035 & 0.3295 & 0.9854 & 0.8862 & 0.9332 & 0.2965 & 0.1138 \\
\bottomrule
\end{tabular}
}

\begin{minipage}{0.98\linewidth}
\footnotesize
\textit{Note:} Class 0 is closed for MRL and blink for RT-BENE; class 1 is
open. All RT-BENE values use the mean of the left- and right-eye open
probabilities before thresholding.
\end{minipage}
\end{table}

\newpage
\subsection{Corruption robustness and modality control}

The complete MRL family/severity summary is given in
Table~\ref{tab:supp_mrl_corruption_results}. It preserves the direction of
both class errors and the participant-cluster interval for Macro-F1.

\scriptsize
\setlength{\tabcolsep}{3pt}
\begin{longtable}{lllcclll}
\caption{MRL corruption-family results under the frozen subject-disjoint test
protocol. Gaussian-noise rows pool five deterministic realizations within
participant and severity. Macro-F1 values include 95\% participant-cluster
bootstrap intervals.}
\label{tab:supp_mrl_corruption_results}\\
\toprule
Model & Perturbation & Severity & Conditions &
Macro-F1 [95\% CI] & BAcc & Closed$\rightarrow$open &
Open$\rightarrow$closed \\
\midrule
\endfirsthead
\toprule
Model & Perturbation & Severity & Conditions &
Macro-F1 [95\% CI] & BAcc & Closed$\rightarrow$open &
Open$\rightarrow$closed \\
\midrule
\endhead
\midrule
\multicolumn{8}{r}{Continued on next page}\\
\endfoot
\bottomrule
\endlastfoot
MobileNetV3-Large & clean & clean & 1 & 0.9794 [0.9704, 0.9963] & 0.9777 & 0.0052 & 0.0393 \\
MobileNetV3-Large & brightness & mild & 1 & 0.9793 [0.9643, 0.9934] & 0.9853 & 0.0086 & 0.0208 \\
MobileNetV3-Large & brightness & moderate & 1 & 0.9807 [0.9720, 0.9951] & 0.9845 & 0.0072 & 0.0237 \\
MobileNetV3-Large & brightness & severe & 1 & 0.9786 [0.9668, 0.9981] & 0.9739 & 0.0041 & 0.0482 \\
MobileNetV3-Large & blur & mild & 1 & 0.9701 [0.9363, 0.9925] & 0.9692 & 0.0082 & 0.0534 \\
MobileNetV3-Large & blur & moderate & 1 & 0.9377 [0.8256, 0.9920] & 0.9128 & 0.0054 & 0.1690 \\
MobileNetV3-Large & blur & severe & 1 & 0.8705 [0.7847, 0.9527] & 0.8206 & 0.0045 & 0.3543 \\
MobileNetV3-Large & noise & mild & 5 & 0.4612 [0.4423, 0.4827] & 0.5000 & 0.0000 & 1.0000 \\
MobileNetV3-Large & noise & moderate & 5 & 0.5190 [0.4790, 0.5937] & 0.5232 & 0.1802 & 0.7733 \\
MobileNetV3-Large & noise & severe & 5 & 0.3896 [0.2860, 0.4426] & 0.3931 & 0.4428 & 0.7709 \\
ShuffleNetV2 x1.0 & clean & clean & 1 & 0.9566 [0.9344, 0.9944] & 0.9417 & 0.0055 & 0.1112 \\
ShuffleNetV2 x1.0 & brightness & mild & 1 & 0.9661 [0.9376, 0.9920] & 0.9638 & 0.0087 & 0.0638 \\
ShuffleNetV2 x1.0 & brightness & moderate & 1 & 0.9669 [0.9472, 0.9931] & 0.9607 & 0.0067 & 0.0719 \\
ShuffleNetV2 x1.0 & brightness & severe & 1 & 0.9564 [0.9346, 0.9981] & 0.9374 & 0.0036 & 0.1216 \\
ShuffleNetV2 x1.0 & blur & mild & 1 & 0.9664 [0.9485, 0.9934] & 0.9599 & 0.0067 & 0.0734 \\
ShuffleNetV2 x1.0 & blur & moderate & 1 & 0.9360 [0.8699, 0.9893] & 0.9058 & 0.0031 & 0.1853 \\
ShuffleNetV2 x1.0 & blur & severe & 1 & 0.8719 [0.8002, 0.9572] & 0.8174 & 0.0012 & 0.3640 \\
ShuffleNetV2 x1.0 & noise & mild & 5 & 0.4612 [0.4423, 0.4827] & 0.5000 & 0.0000 & 1.0000 \\
ShuffleNetV2 x1.0 & noise & moderate & 5 & 0.4612 [0.4421, 0.4825] & 0.4995 & 0.0014 & 0.9997 \\
ShuffleNetV2 x1.0 & noise & severe & 5 & 0.4657 [0.4514, 0.5078] & 0.4787 & 0.0772 & 0.9655 \\
EfficientNet-B0 & clean & clean & 1 & 0.9625 [0.9427, 0.9943] & 0.9533 & 0.0066 & 0.0867 \\
EfficientNet-B0 & brightness & mild & 1 & 0.9617 [0.9012, 0.9876] & 0.9546 & 0.0077 & 0.0830 \\
EfficientNet-B0 & brightness & moderate & 1 & 0.9613 [0.9398, 0.9954] & 0.9512 & 0.0065 & 0.0912 \\
EfficientNet-B0 & brightness & severe & 1 & 0.9321 [0.8515, 0.9951] & 0.9394 & 0.0233 & 0.0979 \\
EfficientNet-B0 & blur & mild & 1 & 0.9689 [0.9484, 0.9923] & 0.9870 & 0.0171 & 0.0089 \\
EfficientNet-B0 & blur & moderate & 1 & 0.9677 [0.9242, 0.9904] & 0.9727 & 0.0116 & 0.0430 \\
EfficientNet-B0 & blur & severe & 1 & 0.9313 [0.8282, 0.9675] & 0.9056 & 0.0065 & 0.1824 \\
EfficientNet-B0 & noise & mild & 5 & 0.2771 [0.2107, 0.3310] & 0.5015 & 0.8116 & 0.1855 \\
EfficientNet-B0 & noise & moderate & 5 & 0.2568 [0.1995, 0.3030] & 0.4970 & 0.8399 & 0.1660 \\
EfficientNet-B0 & noise & severe & 5 & 0.2349 [0.1721, 0.2728] & 0.5099 & 0.8756 & 0.1045 \\
DeiT-Tiny & clean & clean & 1 & 0.9752 [0.9657, 0.9941] & 0.9844 & 0.0112 & 0.0200 \\
DeiT-Tiny & brightness & mild & 1 & 0.9575 [0.9348, 0.9924] & 0.9449 & 0.0064 & 0.1038 \\
DeiT-Tiny & brightness & moderate & 1 & 0.9716 [0.9586, 0.9951] & 0.9709 & 0.0078 & 0.0504 \\
DeiT-Tiny & brightness & severe & 1 & 0.9276 [0.7836, 0.9945] & 0.9721 & 0.0438 & 0.0119 \\
DeiT-Tiny & blur & mild & 1 & 0.9727 [0.9616, 0.9952] & 0.9800 & 0.0111 & 0.0289 \\
DeiT-Tiny & blur & moderate & 1 & 0.9612 [0.9422, 0.9950] & 0.9554 & 0.0085 & 0.0808 \\
DeiT-Tiny & blur & severe & 1 & 0.9260 [0.8411, 0.9820] & 0.8931 & 0.0040 & 0.2098 \\
DeiT-Tiny & noise & mild & 5 & 0.9690 [0.9552, 0.9930] & 0.9719 & 0.0102 & 0.0460 \\
DeiT-Tiny & noise & moderate & 5 & 0.9155 [0.8681, 0.9587] & 0.9288 & 0.0316 & 0.1107 \\
DeiT-Tiny & noise & severe & 5 & 0.8090 [0.6985, 0.8774] & 0.8746 & 0.1073 & 0.1435 \\
RepViT-M1.0 & clean & clean & 1 & 0.9631 [0.9441, 0.9954] & 0.9538 & 0.0064 & 0.0860 \\
RepViT-M1.0 & brightness & mild & 1 & 0.9539 [0.9090, 0.9898] & 0.9460 & 0.0095 & 0.0986 \\
RepViT-M1.0 & brightness & moderate & 1 & 0.9669 [0.9500, 0.9951] & 0.9607 & 0.0067 & 0.0719 \\
RepViT-M1.0 & brightness & severe & 1 & 0.9364 [0.8753, 0.9669] & 0.9621 & 0.0313 & 0.0445 \\
RepViT-M1.0 & blur & mild & 1 & 0.9530 [0.9221, 0.9898] & 0.9448 & 0.0096 & 0.1008 \\
RepViT-M1.0 & blur & moderate & 1 & 0.9150 [0.8530, 0.9769] & 0.8854 & 0.0082 & 0.2209 \\
RepViT-M1.0 & blur & severe & 1 & 0.8683 [0.7937, 0.9637] & 0.8232 & 0.0081 & 0.3454 \\
RepViT-M1.0 & noise & mild & 5 & 0.6556 [0.5632, 0.7681] & 0.6144 & 0.0002 & 0.7711 \\
RepViT-M1.0 & noise & moderate & 5 & 0.4709 [0.4558, 0.5023] & 0.5012 & 0.0102 & 0.9874 \\
RepViT-M1.0 & noise & severe & 5 & 0.5693 [0.5234, 0.6499] & 0.5596 & 0.0572 & 0.8236 \\
EfficientFormer-L1 & clean & clean & 1 & 0.9780 [0.9579, 0.9931] & 0.9834 & 0.0087 & 0.0245 \\
EfficientFormer-L1 & brightness & mild & 1 & 0.8521 [0.6229, 0.9599] & 0.9394 & 0.1011 & 0.0200 \\
EfficientFormer-L1 & brightness & moderate & 1 & 0.9595 [0.8641, 0.9928] & 0.9765 & 0.0196 & 0.0274 \\
EfficientFormer-L1 & brightness & severe & 1 & 0.9825 [0.9679, 0.9921] & 0.9884 & 0.0076 & 0.0156 \\
EfficientFormer-L1 & blur & mild & 1 & 0.9719 [0.9325, 0.9908] & 0.9794 & 0.0115 & 0.0297 \\
EfficientFormer-L1 & blur & moderate & 1 & 0.9507 [0.8483, 0.9852] & 0.9396 & 0.0088 & 0.1119 \\
EfficientFormer-L1 & blur & severe & 1 & 0.9158 [0.8013, 0.9725] & 0.8813 & 0.0054 & 0.2320 \\
EfficientFormer-L1 & noise & mild & 5 & 0.5211 [0.3694, 0.6394] & 0.6245 & 0.4043 & 0.3466 \\
EfficientFormer-L1 & noise & moderate & 5 & 0.1258 [0.0628, 0.1715] & 0.5000 & 1.0000 & 0.0000 \\
EfficientFormer-L1 & noise & severe & 5 & 0.3679 [0.2785, 0.4175] & 0.3785 & 0.5021 & 0.7408 \\
\end{longtable}

The worst target-domain conditions determining the primary safety-related
screen are listed in Table~\ref{tab:supp_safety_diagnostics}. These values explain why latency compliance did not imply operational eligibility.
\vspace{-0.4cm}

\begin{table}[!h]
\centering
\caption{Condition-level diagnostics supporting the primary non-compensatory safety-related screen. Values are computed over the frozen non-clean RT-BENE out-of-fold conditions. The directional-error ceiling of 0.50 is equivalent to requiring at least 0.50 recall for each class under every evaluated condition; class collapse is not permitted.}
\label{tab:supp_safety_diagnostics}
\scriptsize
\setlength{\tabcolsep}{3pt}
\resizebox{\linewidth}{!}{%
\begin{tabular}{llrlrrrr}
\toprule
Model & Worst blink$\rightarrow$open condition &
Blink$\rightarrow$open & Worst open$\rightarrow$blink condition &
Open$\rightarrow$blink & Worst Macro-F1 & Worst BAcc & Collapse count \\
\midrule
MobileNetV3-Large & noise\_moderate\_r05 & 0.2293 & noise\_severe\_r01 & 0.8177 & 0.1897 & 0.5061 & 0 \\
ShuffleNetV2 x1.0 & brightness\_moderate & 0.1788 & noise\_severe\_r01 & 1.0000 & 0.0361 & 0.5000 & 10 \\
EfficientNet-B0 & noise\_severe\_r04 & 0.8382 & noise\_mild\_r02 & 0.8302 & 0.1878 & 0.4566 & 0 \\
DeiT-Tiny & noise\_severe\_r02 & 0.6947 & blur\_severe & 0.3105 & 0.4896 & 0.6178 & 0 \\
RepViT-M1.0 & noise\_moderate\_r02 & 1.0000 & blur\_severe & 0.3948 & 0.4353 & 0.4991 & 15 \\
EfficientFormer-L1 & noise\_severe\_r05 & 0.7071 & noise\_moderate\_r02 & 0.9940 & 0.0421 & 0.4636 & 5 \\
\bottomrule
\end{tabular}
}
\end{table}

The Gray3 chroma-ablation results are summarized in
Table~\ref{tab:supp_gray3_summary}. The control changed performance
substantially for several architectures, but does not isolate sensing modality
or reproduce near-infrared image formation.
\vspace{-0.4cm}
\begin{table}[!h]
\centering
\caption{Three-channel grayscale (Gray3) chroma-ablation summary on the
matched 11-participant RT-BENE cohort. Changes are Gray3 minus RGB and include
95\% paired participant-cluster bootstrap intervals. Gray3 is not interpreted
as a simulation of near-infrared sensing.}
\label{tab:supp_gray3_summary}
\tiny
\setlength{\tabcolsep}{2.5pt}
\resizebox{\linewidth}{!}{%
\begin{tabular}{lrrcrrcrr}
\toprule
& \multicolumn{3}{c}{Clean Macro-F1} &
\multicolumn{3}{c}{Mean non-clean family Macro-F1} &
\multicolumn{2}{c}{Worst-family Macro-F1}\\
\cmidrule(lr){2-4}\cmidrule(lr){5-7}\cmidrule(lr){8-9}
Model & RGB & Gray3 & $\Delta$ [95\% CI] &
RGB & Gray3 & $\Delta$ [95\% CI] & RGB & Gray3 \\
\midrule
MobileNetV3-Large & 0.5474 & 0.7756 & 0.2282 [0.0917, 0.2997] & 0.3405 & 0.5299 & 0.1894 [0.1353, 0.2313] & 0.0362 & 0.0468 \\
ShuffleNetV2 x1.0 & 0.2323 & 0.6695 & 0.4372 [0.3171, 0.6212] & 0.2298 & 0.5255 & 0.2956 [0.2063, 0.3946] & 0.0391 & 0.1231 \\
EfficientNet-B0 & 0.5001 & 0.6523 & 0.1522 [0.0584, 0.3273] & 0.4214 & 0.5025 & 0.0810 [-0.0224, 0.1882] & 0.2133 & 0.1908 \\
DeiT-Tiny & 0.7443 & 0.7585 & 0.0142 [0.0004, 0.0278] & 0.6756 & 0.6800 & 0.0045 [-0.0100, 0.0197] & 0.5329 & 0.5408 \\
RepViT-M1.0 & 0.5797 & 0.7161 & 0.1364 [0.0099, 0.2288] & 0.4934 & 0.5696 & 0.0762 [0.0417, 0.1028] & 0.3851 & 0.3404 \\
EfficientFormer-L1 & 0.5928 & 0.7172 & 0.1244 [0.0852, 0.1450] & 0.5108 & 0.6070 & 0.0962 [0.0788, 0.1164] & 0.3636 & 0.4912 \\
\bottomrule
\end{tabular}
}
\end{table}

\newpage
\subsection{Matched target-domain effects}

The primary target-domain protocol used three participant-safe outer folds. Three participants were reserved as a fixed validation set, while the remaining 11 participants were assigned to mutually exclusive outer test folds containing four, four, and three participants. All sessions belonging to the same participant were kept within one role in each fold, so no participant contributed simultaneously to training, validation, and testing. Each of the six architectures was trained independently in each outer fold from fresh ImageNet-1K initialization, yielding 18 model-fold runs and one out-of-fold prediction for every frame in the 77,119-frame matched cohort.

Table~\ref{tab:supp_adaptation_effects} reports the complete matched B1--B2
comparison. Positive changes favor B2 only for higher-is-better metrics;
negative changes favor B2 for Brier and the directional error rates.

\begin{table}[!h]
\centering
\caption{Matched zero-shot (B1) and participant-safe target-domain out-of-fold
(B2) effects on the same 11 participants and 77,119 RT-BENE frames. Changes
are B2 minus B1; intervals are 95\% paired participant-cluster bootstrap
intervals.}
\label{tab:supp_adaptation_effects}
\scriptsize
\setlength{\tabcolsep}{2pt}

\textbf{Panel A: discrimination metrics}\\[2mm]
\resizebox{\linewidth}{!}{%
\begin{tabular}{lrrcrrcrrc}
\toprule
& \multicolumn{3}{c}{Macro-F1} &
\multicolumn{3}{c}{Balanced accuracy} &
\multicolumn{3}{c}{AUROC}\\
\cmidrule(lr){2-4}\cmidrule(lr){5-7}\cmidrule(lr){8-10}
Model & B1 & B2 & $\Delta$ [95\% CI] &
B1 & B2 & $\Delta$ [95\% CI] &
B1 & B2 & $\Delta$ [95\% CI] \\
\midrule
MobileNetV3-Large & 0.5474 & 0.6464 & 0.0990 [0.0263, 0.2522] & 0.8520 & 0.8977 & 0.0457 [-0.0027, 0.1181] & 0.8668 & 0.9457 & 0.0789 [0.0082, 0.1816] \\
ShuffleNetV2 x1.0 & 0.2323 & 0.7130 & 0.4807 [0.3794, 0.6135] & 0.6071 & 0.9019 & 0.2948 [0.2069, 0.3614] & 0.6558 & 0.9647 & 0.3089 [0.2134, 0.4080] \\
EfficientNet-B0 & 0.5001 & 0.6503 & 0.1502 [0.0458, 0.3329] & 0.7913 & 0.8672 & 0.0759 [-0.0163, 0.1715] & 0.8402 & 0.9428 & 0.1026 [0.0193, 0.2222] \\
DeiT-Tiny & 0.7443 & 0.7037 & -0.0406 [-0.0890, 0.0651] & 0.9103 & 0.8762 & -0.0341 [-0.0646, 0.0073] & 0.9696 & 0.9637 & -0.0058 [-0.0346, 0.0184] \\
RepViT-M1.0 & 0.5797 & 0.5667 & -0.0130 [-0.0989, 0.0212] & 0.7923 & 0.7479 & -0.0444 [-0.1543, 0.0223] & 0.9089 & 0.8626 & -0.0462 [-0.1319, 0.0000] \\
EfficientFormer-L1 & 0.5928 & 0.6390 & 0.0462 [-0.0290, 0.1725] & 0.7763 & 0.8275 & 0.0512 [-0.0172, 0.0765] & 0.8867 & 0.9422 & 0.0556 [0.0121, 0.0901] \\
\bottomrule
\end{tabular}
}

\vspace{5mm}
\textbf{Panel B: probability quality and directional errors}\\[2mm]
\resizebox{\linewidth}{!}{%
\begin{tabular}{lrrcrrcrrc}
\toprule
& \multicolumn{3}{c}{Brier score} &
\multicolumn{3}{c}{Blink$\rightarrow$open} &
\multicolumn{3}{c}{Open$\rightarrow$blink}\\
\cmidrule(lr){2-4}\cmidrule(lr){5-7}\cmidrule(lr){8-10}
Model & B1 & B2 & $\Delta$ [95\% CI] &
B1 & B2 & $\Delta$ [95\% CI] &
B1 & B2 & $\Delta$ [95\% CI] \\
\midrule
MobileNetV3-Large & 0.1986 & 0.0950 & -0.1036 [-0.2297, -0.0206] & 0.0553 & 0.0806 & 0.0252 [0.0114, 0.0435] & 0.2406 & 0.1239 & -0.1167 [-0.2734, -0.0168] \\
ShuffleNetV2 x1.0 & 0.5772 & 0.0529 & -0.5243 [-0.6553, -0.4025] & 0.0159 & 0.1228 & 0.1068 [0.0377, 0.2691] & 0.7698 & 0.0734 & -0.6964 [-0.8235, -0.5783] \\
EfficientNet-B0 & 0.1912 & 0.0830 & -0.1083 [-0.2160, -0.0350] & 0.1120 & 0.1556 & 0.0436 [-0.0637, 0.2442] & 0.3054 & 0.1100 & -0.1954 [-0.3996, -0.0646] \\
DeiT-Tiny & 0.0458 & 0.0569 & 0.0111 [-0.0223, 0.0579] & 0.1210 & 0.1753 & 0.0543 [-0.0143, 0.0976] & 0.0583 & 0.0723 & 0.0140 [-0.0278, 0.0676] \\
RepViT-M1.0 & 0.1098 & 0.1231 & 0.0134 [-0.0243, 0.0533] & 0.2531 & 0.3437 & 0.0906 [-0.1036, 0.3775] & 0.1623 & 0.1605 & -0.0018 [-0.0778, 0.0743] \\
EfficientFormer-L1 & 0.1097 & 0.0651 & -0.0446 [-0.1015, 0.0082] & 0.3105 & 0.2386 & -0.0719 [-0.1746, 0.1062] & 0.1369 & 0.1064 & -0.0304 [-0.0876, 0.0552] \\
\bottomrule
\end{tabular}
}
\end{table}

\end{landscape}

\subsection{Embedded rank transfer}

\begin{figure}[!h]
\centering
\includegraphics[width=0.7\textwidth]{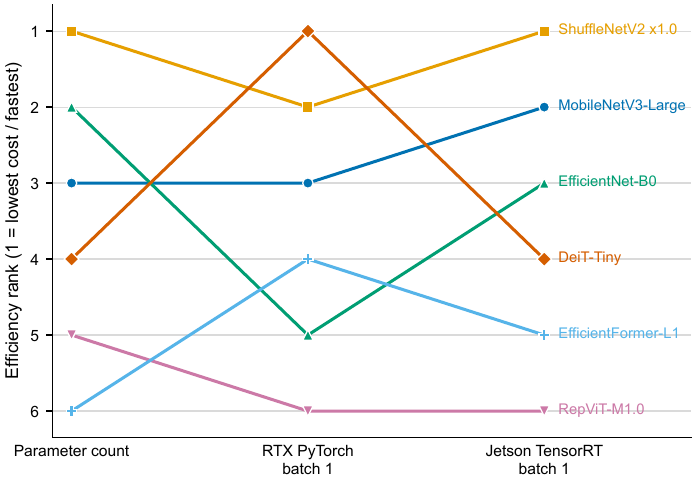}
\caption{Efficiency-rank transfer across parameter count, RTX PyTorch
batch-1 latency, and Jetson TensorRT batch-1 latency. Rank 1 denotes the
lowest cost or fastest execution within each column.}
\label{fig:supp_efficiency_rank_transfer}
\end{figure}

The complete rank transfer from parameter count to RTX PyTorch batch-1 timing
and Jetson TensorRT batch-1 timing is shown in
Fig.~\ref{fig:supp_efficiency_rank_transfer}. The result illustrates why a
static-complexity ranking is not interchangeable with a target-stack ranking.

All 12 TensorRT FP32 engines passed the prespecified numerical-parity gate before latency measurement. Parity was evaluated on 16 frozen real-image MRL Eye target vectors against the frozen reference outputs. Acceptance required $P99(|\Delta p|)\leq10^{-4}$, $\max|\Delta p|\leq5\times10^{-4}$, zero class disagreements, and a maximum repeated-run absolute logit difference no greater than $10^{-7}$. This gate verifies numerical consistency on the frozen validation inputs and is not a full-test-set accuracy-equivalence evaluation.

\subsection{RISE insertion coverage and stochastic stability}
Pairwise subject-level inference covered five prespecified faithfulness
metrics: normalized deletion AUC, raw deletion AUC, raw insertion AUC,
normalized insertion AUC, and a derived explanation score. The latter was
defined as $0.5[(1-D_{\mathrm{norm}})+I_{\mathrm{norm}}]$, with higher
values preferred. The first three metrics were evaluated on all 500
common-correct images per model, whereas normalized insertion and the
derived score were compared only on the 49-image six-model common-valid
intersection. With six models, the 15 pairwise comparisons for each metric
yielded 75 exact tests in total. Holm adjustment was applied separately within the 15 model-pair comparisons
of each metric. Because only five participants were available, the minimum
attainable two-sided exact $p$-value was 0.0625; the resulting inference was
therefore treated as exploratory.

\begin{figure}[!h]
\centering
\includegraphics[width=0.82\textwidth]{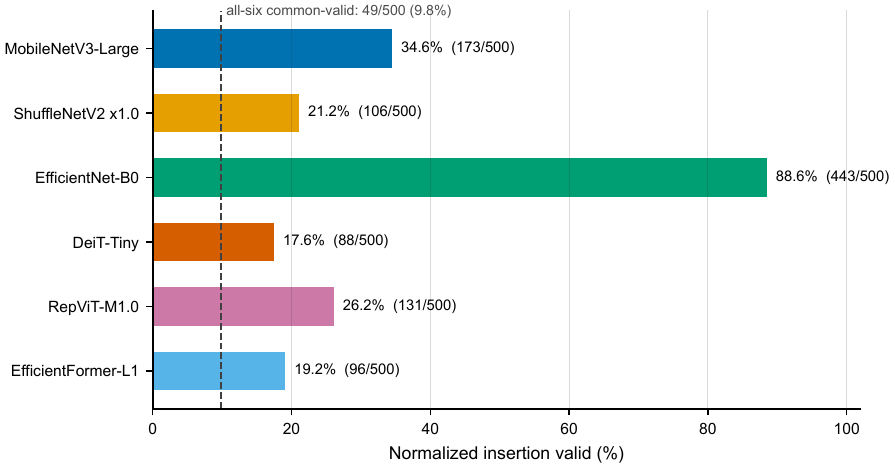}
\caption{Model-specific validity of normalized insertion under
$|f_0-f_b|\geq0.05$. Labels show the percentage and number of valid images
out of 500.}
\label{fig:supp_insertion_coverage}
\end{figure}

Normalized insertion was defined only when the input--baseline target-score
contrast satisfied the frozen denominator requirement. Model-specific
coverage is shown in Fig.~\ref{fig:supp_insertion_coverage}. Coverage is a
metric-availability diagnostic, not a faithfulness score.

\begin{figure}[!h]
\centering
\includegraphics[width=0.82\textwidth]{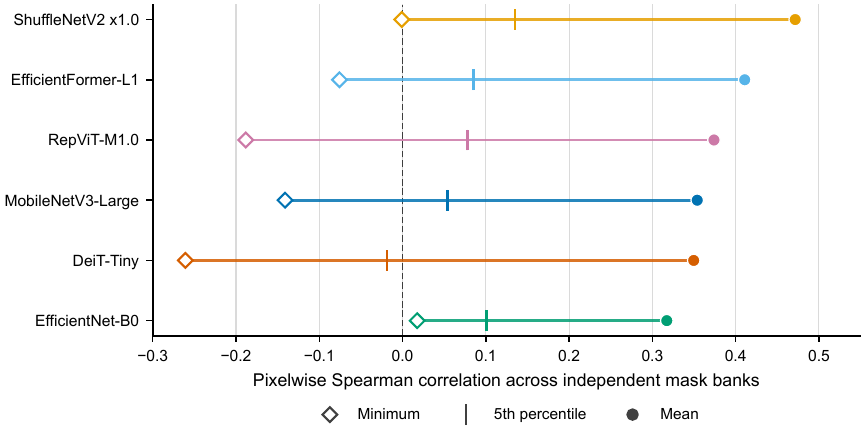}
\caption{Pixelwise Spearman correlation across independent RISE mask banks.
Filled circles show the model-specific mean, vertical ticks the fifth
percentile, and open diamonds the minimum observed value.}
\label{fig:supp_spearman_stability}
\end{figure}

\begin{figure}[!h]
\centering
\includegraphics[width=0.82\textwidth]{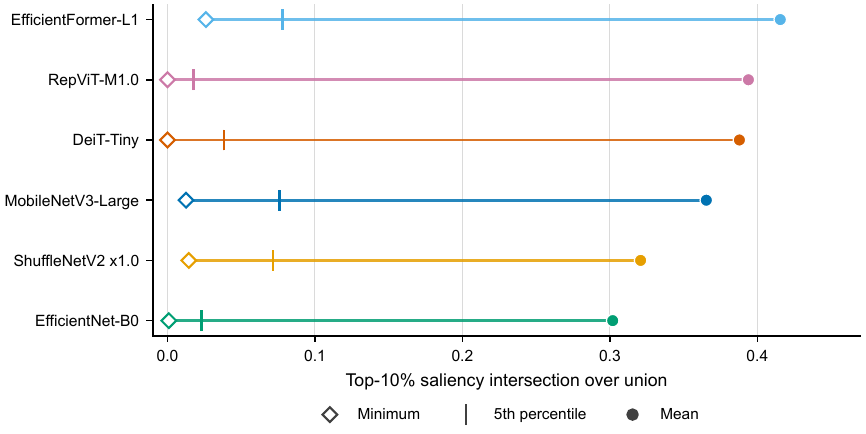}
\caption{Top-10\% saliency intersection over union across independent RISE
mask banks. Filled circles show the model-specific mean, vertical ticks the
fifth percentile, and open diamonds the minimum observed value.}
\label{fig:supp_iou_stability}
\end{figure}

Figures~\ref{fig:supp_spearman_stability} and
\ref{fig:supp_iou_stability} report the stochastic stability of RISE maps
across three independent 2,048-mask banks on the balanced 64-image subset.
The mean summarizes all pairwise bank comparisons; the minimum shows the
least stable observed comparison.

\subsection{Historical HCS sensitivity}

The original HCS profiles and global preference-space summaries are reported
in Table~\ref{tab:supp_hcs_sensitivity}. The predefined profiles reproduce
the historical ranking, whereas the exhaustive audit shows that the full
ordering remains weight dependent.

\begin{landscape}
\begin{table}[!h]
\centering
\caption{Retrospective historical HCS audit. Values in parentheses are ranks
under the three original profiles. First-place and rank shares quantify
coverage of the audited weight space and are not probabilities of objective
model superiority. RISE and the two added architectures are excluded.}
\label{tab:supp_hcs_sensitivity}
\scriptsize
\setlength{\tabcolsep}{2.5pt}
\resizebox{\linewidth}{!}{%
\begin{tabular}{lcccrrrrrrr}
\toprule
Model & Safety HCS & Deployment HCS & Balanced HCS &
Lattice first (\%) & Dirichlet first (\%) &
Rank 1 (\%) & Rank 2 (\%) & Rank 3 (\%) & Rank 4 (\%) & Mean rank \\
\midrule
MobileNetV3-Large & 0.7920 (2) & 0.7687 (2) & 0.7659 (2) & 0.064 & 0.058 & 0.064 & 57.030 & 25.248 & 17.657 & 2.605 \\
ShuffleNetV2 x1.0 & 0.8604 (1) & 0.8795 (1) & 0.8508 (1) & 74.568 & 75.489 & 74.568 & 18.491 & 6.752 & 0.188 & 1.326 \\
EfficientNet-B0 & 0.7407 (4) & 0.6698 (3) & 0.7096 (4) & 10.926 & 10.521 & 10.926 & 11.738 & 38.283 & 39.053 & 3.055 \\
DeiT-Tiny & 0.7625 (3) & 0.6603 (4) & 0.7099 (3) & 14.441 & 13.932 & 14.441 & 12.740 & 29.718 & 43.101 & 3.015 \\
\bottomrule
\end{tabular}
}
\end{table}
\end{landscape}

\begin{figure}[!h]
\centering
\includegraphics[width=0.92\textwidth]{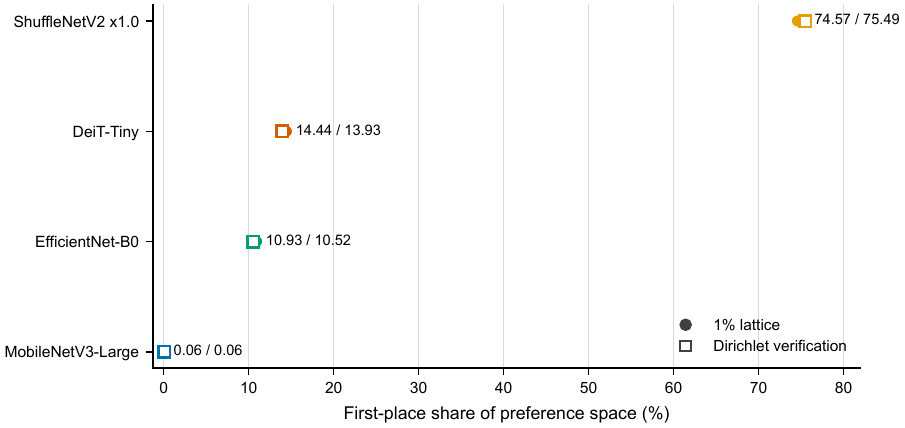}
\caption{First-place allocation over the historical HCS preference space.
The exhaustive 1\% lattice and the deterministic Dirichlet verification
produce closely aligned allocations.}
\label{fig:supp_hcs_first_place}
\end{figure}

\begin{figure}[!h]
\centering
\includegraphics[width=0.92\textwidth]{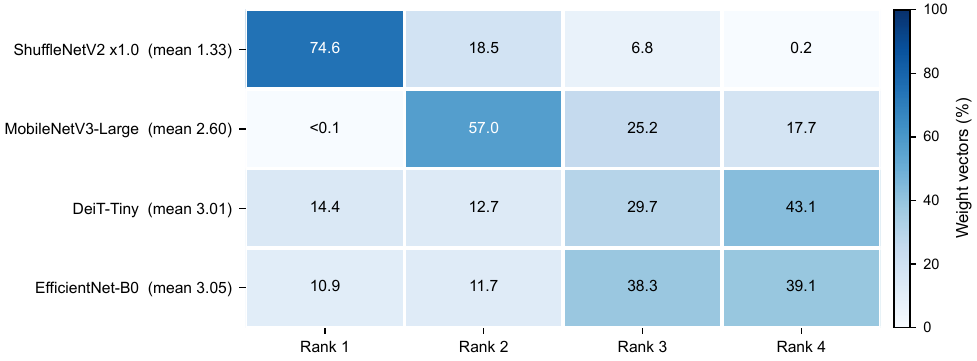}
\caption{Competition-rank distribution over the exhaustive historical HCS
weight lattice. All four ranks occur for multiple models, and all 24 strict
orderings occur somewhere in the preference space.}
\label{fig:supp_hcs_rank_distribution}
\end{figure}

Figure~\ref{fig:supp_hcs_first_place} compares first-place allocation under
the exhaustive lattice and Dirichlet verification. Figure~
\ref{fig:supp_hcs_rank_distribution} shows the complete lattice rank
distribution for each historical model.

\subsection{Extended qualitative RISE gallery}
\label{subsec:supp_rise_qualitative}

\begin{landscape}
\begin{figure}[!h]
    \centering
    \includegraphics[width=0.98\linewidth]
    {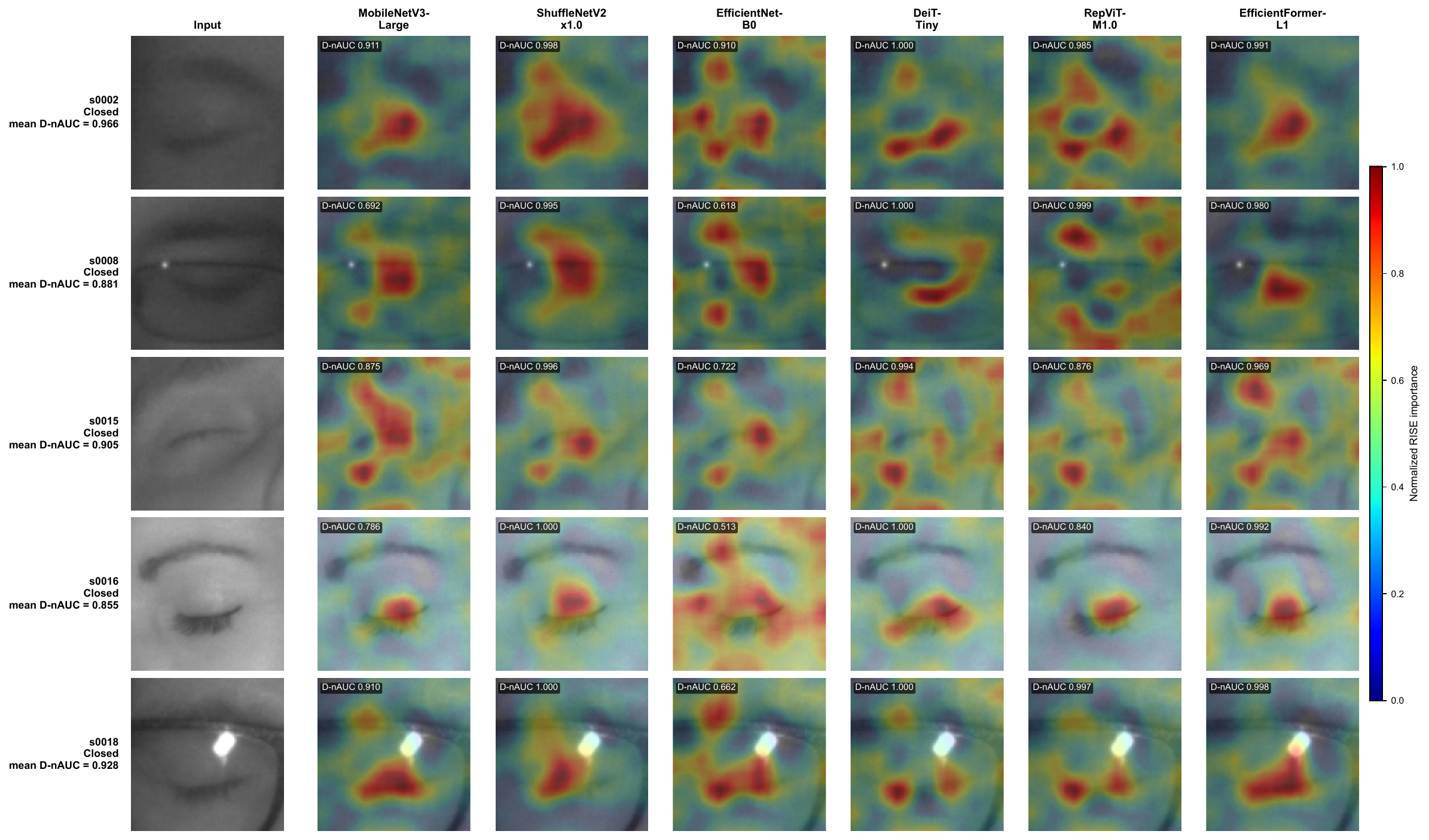}
    \caption{Subject-stratified qualitative RISE comparison for closed-eye
    images. One common-correct image was selected for each of the five MRL
    test participants as the sample closest to the corresponding
    participant--class median of the six-model mean normalized deletion AUC.
    Values displayed on the maps report normalized deletion AUC (D-nAUC), for
    which lower values are preferred. All maps explain the ground-truth class
    and use a common visualization scale.}
    \label{fig:supp_rise_closed_gallery}
\end{figure}
\end{landscape}

\begin{landscape}
\begin{figure}[!h]
    \centering
    \includegraphics[width=0.98\linewidth]
    {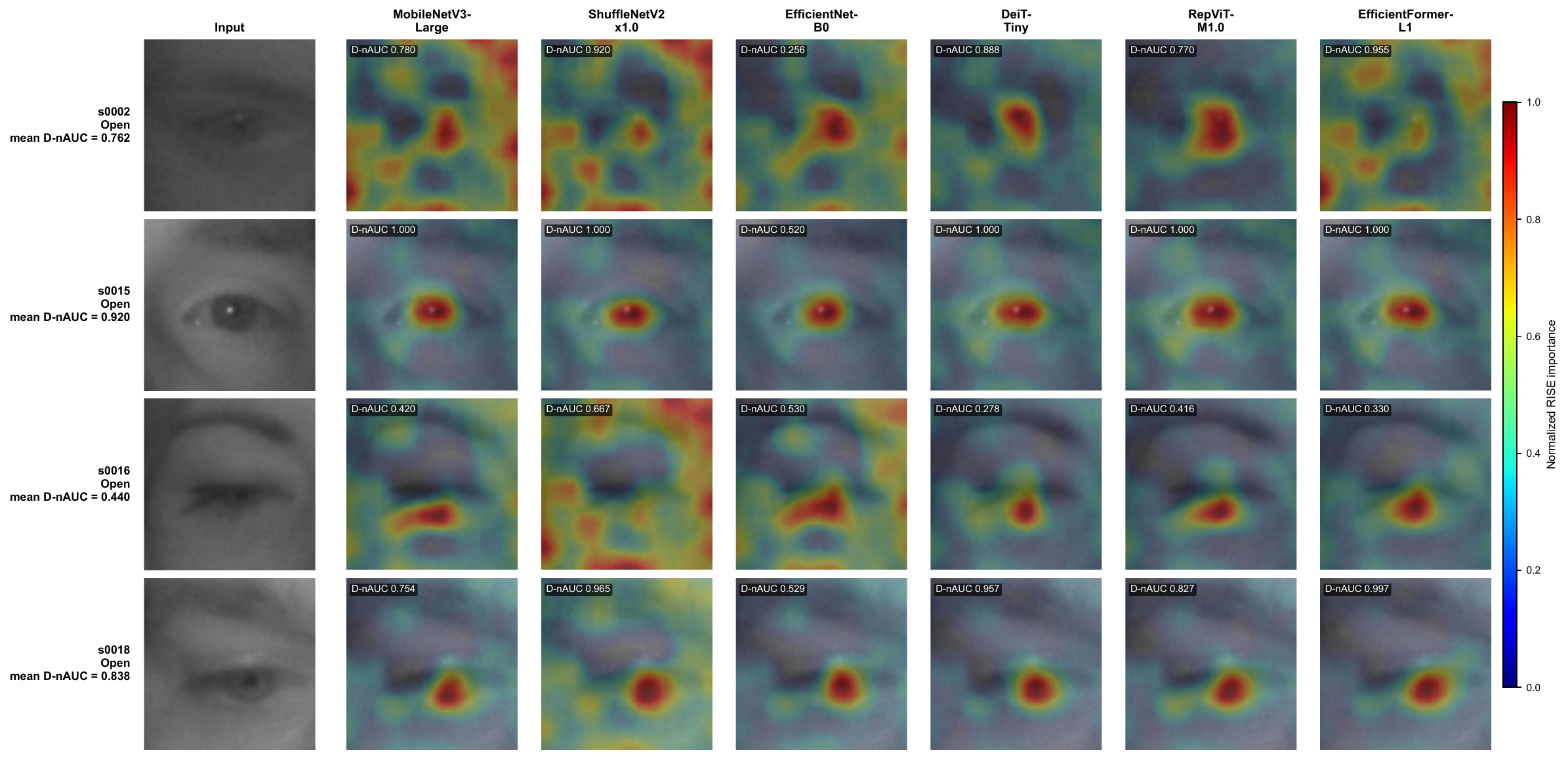}
    \caption{Subject-stratified qualitative RISE comparison for open-eye
    images. One common-correct image was selected for each participant with
    an eligible open-eye sample in the frozen XAI corpus using the same
    median-based rule as Fig.~\ref{fig:supp_rise_closed_gallery}. Participant
    s0008 is not shown because no eligible open-eye sample was available.
    Values displayed on the maps report normalized deletion AUC (D-nAUC), for
    which lower values are preferred.}
    \label{fig:supp_rise_open_gallery}
\end{figure}
\end{landscape}

The two examples reported in the main manuscript were complemented by a
subject-stratified qualitative gallery. Within each available participant--class
stratum, the selected common-correct image was the sample whose six-model mean
normalized deletion AUC was closest to the corresponding stratum median.
Selection was performed without visual inspection. Closed- and open-eye
examples are shown in Figs.~\ref{fig:supp_rise_closed_gallery} and
\ref{fig:supp_rise_open_gallery}, respectively. All maps use the ground-truth
class as the explanation target, the frozen RISE configuration, and a common
visualization scale. Participant s0008 had no eligible open-eye sample in the
frozen XAI corpus and is therefore absent from
Fig.~\ref{fig:supp_rise_open_gallery}. The galleries are illustrative and do
not replace the subject-level quantitative comparison.

